\documentclass[a4paper,final]{article}
\usepackage[dvips]{graphicx}
\usepackage{amsmath} % Extra environments for multiline displayed eq
\usepackage{amssymb} % Defines symbol names 
\renewenvironment{abstract}
  {\normalfont
    \list{}{\labelwidth0pt
      \leftmargin0pt \rightmargin\leftmargin
      \listparindent\parindent \itemindent0pt
      \parsep0pt
      
    }
    \item[\hskip\labelsep\bfseries\abstractname\enspace --] \itshape
}{
  \endlist}

\newcommand{\keywordsname}{Keywords}
\newenvironment{keywords}
  {\normalfont
    \list{}{\labelwidth0pt
      \leftmargin0pt \rightmargin\leftmargin
      \listparindent\parindent \itemindent0pt
      \parsep0pt
      }
    \item[\hskip\labelsep\bfseries\keywordsname:]}{\endlist}

\begin{document}

\pagestyle{myheadings}
\markboth{}{}

\title{Belief Conditioning Rules}
\author{
Florentin Smarandache\\
Department of Mathematics,\\
University of New Mexico,\\
Gallup, NM 87301, U.S.A.\\
smarand@unm.edu\\
\and
Jean Dezert\\
ONERA,\\
29 Avenue de la  Division Leclerc, \\
92320 Ch\^{a}tillon, France.\\
Jean.Dezert@onera.fr}

\date{}

\maketitle
\vspace{2cm}

\begin{abstract}
In this paper we propose a new family of Belief Conditioning Rules (BCR)\index{BCR (Belief Conditioning Rule)} for belief revision. These rules are not directly related with the fusion of several sources of evidence but with the revision of a belief assignment available at a given time according to the new truth (i.e. conditioning constraint) one has about the space of solutions of the problem. 
\end{abstract}


% ================
\section{Introduction}
% ================
\label{BCRintro}

In this paper we define several Belief Conditioning Rules (BCR) for use in information fusion and for belief revision. Suppose we have a basic belief assignment (bba) $m_1(.)$ defined on hyper-power set $D^\Theta$, and we find out that the truth is in a given element $A\in D^\Theta$. So far in literature devoted to belief functions and the mathematical theory of evidence, there has been used Shafer's Conditioning Rule (SCR) \cite{Shafer_1976}, which simply combines the mass $m_1(.)$ with a specific bba focused on $A$, i.e. $m_S(A)=1$, and then uses Dempster's rule to transfer the conflicting mass to non-empty sets. But in our opinion this conditioning approach based on the combination of two bba's is {\it{subjective}} since in such procedure both sources are subjective. While conditioning a mass $m_1(.)$, {\it{knowing}} (or assuming) that the truth is in $A$, means that we have an absolute (not subjective) information, i.e. the truth is in $A$ has occurred (or is assumed to have occurred), thus $A$ was realized (or is assumed to be realized), hence it is an absolute truth. "Truth in $A$" must therefore be considered as an absolute truth when conditioning, while $m_S(A)=1$ used in SCR does not refer to an absolute truth actually, but only to a subjective certainty in the possible occurrence of $A$ given by a second source of evidence. This is the main and fundamental distinction between our approaches (BCRs) and Shafer's (SCR).  In our opinion, SCR does not do a conditioning, but only a fusion of $m_1(.)$ with a particular bba $m_S(A)=1$. The main advantage of SCR is that it is simple and thus very appealing, and in some cases it gives the same results with some BCRs, and it remains coherent with conditional probability when $m_1(.)$ is a Bayesian belief assignment. In the sequel, we will present many (actually thirty one BCR rules, denoted BCR1-BCR31) new alternative issues for belief conditioning. The sequel does not count: a) if we first know the source $m_1(.)$ and then that the truth is in $A$ (or is supposed to be in $A$), or b) if we first know (or assume) the truth is in $A$, and then we find the source $m_1()$.The results of conditioning are the same. In addition, we work on a hyper-power set, that is a generalization of the power set. The best among these BCR1-31, that we recommend researchers to use, are: BCR17 for a pessimistic/prudent view on conditioning problem and a more refined redistribution of conflicting masses, or BCR12 for a very pessimistic/prudent view and less refined redistribution. After a short presentation of SCR rule, we present in the following sections all new BCR rules we propose, many examples, and a very important and open challenging question about belief fusion and conditioning.

% ================
\section{Shafer's conditioning rule (SCR)}
% ================

Before going further in the development of new belief conditioning rules\index{BCR (Belief Conditioning Rule)}, it is important to recall the  conditioning of beliefs proposed by Glenn Shafer in \cite{Shafer_1976} (p.66--67) and reported below. \\

So, let's suppose that the effect of a new evidence (say source 2) on the frame of discernment $\Theta$ is to establish a particular subset $B\subset \Theta$ with certainty. Then $\text{Bel}_2$ will give a degree of belief one to the proposition corresponding to $B$ and to every proposition implied by it:
$$\text{Bel}_2(A)=
\begin{cases}
1, \qquad \text{if}\, B\subset A;\\
0, \qquad \text{otherwise.}\\
\end{cases}
$$

Since the  subset $B$ is the only focal element of $\text{Bel}_2$, its basic belief assignment is one, i.e. $m_2(B)=1$. Such a function $\text{Bel}_2$ is then combinable with the (prior) $\text{Bel}_1$ as long as $\text{Bel}_1(\bar{B}) < 1$, and the Dempster's rule\index{Dempster's rule} of combination (denoted $\oplus$) provides the conditional belief $\text{Bel}_1(.|B)=\text{Bel}_1\oplus\text{Bel}_2$ (according to Theorem 3.6 in \cite{Shafer_1976}). More specifically, one gets for all $A\subset\Theta$,

$$\text{Bel}_1(A|B)=
\frac{\text{Bel}_1(A\cup \bar{B})-\text{Bel}_1(\bar{B})}{1-\text{Bel}_1(\bar{B})}
$$

$$\text{Pl}_1(A|B)=
\frac{\text{Pl}_1(A\cap B)}{\text{Pl}_1(B)}
$$

\noindent
where $\text{Pl}(.)$ denotes the plausibility function.

% ================
\section{Belief Conditioning Rules (BCR)}
% ================
\label{BCRrules}

Let $\Theta=\{\theta_1,\theta_2,\ldots,\theta_n\}$, $n\geq 2$, and the hyper-power set\footnote{The below formulas can also be defined on the power set $2^\Theta$ and respectively super-power set $S^\Theta$ in exactly the same way.} $D^\Theta$. Let's consider a basic belief assignment (bba) $m(.): D^\Theta \mapsto [0,1]$ such that $\sum_{X\in D^\Theta}m(X)=1$.\\

Suppose one finds out that the truth is in the set $A\in D^\Theta\setminus\{\emptyset\}$. Let $\mathcal{P}_{\mathcal{D}}(A)=2^A \cap D^{\Theta} \setminus \{\emptyset\}$, i.e. all non-empty parts (subsets) of $A$ which are included in $D^\Theta$. Let's consider the normal 
cases when $A\neq\emptyset$ and $\sum_{Y\in \mathcal{P}_{\mathcal{D}}(A)}m(Y)> 0$. For the degenerate case when the truth is in $A=\emptyset$, we consider Smets' open-world, which means that there are other hypotheses $\Theta'=\{\theta_{n+1},\theta_{n+2},\ldots\theta_{n+m}\}$, $m\geq 1$, and the truth is in $A\in D^{\Theta'}\setminus\{\emptyset\}$. If $A=\emptyset$ and we consider a close-world, then it means that the problem is impossible. For another degenerate case, when $\sum_{Y\in \mathcal{P}_{\mathcal{D}}(A)}m(Y)=0$, i.e. when the source gave us a totally (100\%) wrong information $m(.)$, then, we define: $m(A|A)\triangleq 1$ and, as a consequence, $m(X|A)=0$ for any $X\neq A$.\\

Let $s(A)=\{\theta_{i_1},\theta_{i_2},\ldots,\theta_{i_p}\}$, $1\leq p\leq n$,  be the singletons/atoms that compose $A$ (For example, if $A=\theta_1\cup(\theta_3\cap\theta_4)$ then $s(A)=\{\theta_1,\theta_3,\theta_4\}$.).
We consider three subsets of $D^\Theta \setminus \emptyset$, generated by $A$:
\begin{itemize}
\item
$D_1=\mathcal{P}_{\mathcal{D}}(A)$, the parts of $A$ which are included in the hyper-power set, except the empty set; 
\item
$D_2=\{(\Theta\setminus s(A)),\cup , \cap\} \setminus \{\emptyset\}$, i.e. the sub-hyper-power set generated by $\Theta\setminus s(A)$ under $\cup$ and $\cap$, without the empty set.
\item
$D_3=(D^\Theta\setminus\{\emptyset\}) \setminus (D_1\cup D_2)$; each set from $D_3$ has in its formula singletons from both $s(A)$ and $\Theta\setminus s(A)$ in the case when $\Theta\setminus s(A)$ 
is different from empty set.
\end{itemize}

\noindent
$D_1$, $D_2$ and $D_3$ have no element in common two by two and their union is $D^\Theta\setminus\{\emptyset\}$.\\

\noindent
{\it{Examples of decomposition of $D^\Theta\setminus\{\emptyset\}=D_1\cup D_2\cup D_3$}}: Let's consider $\Theta=\{A, B, C\}$ and the free DSm model\index{Dezert-Smarandache free model}.

\begin{itemize}
\item If one supposes the truth is in $A$, then $D_1 = \{A, A\cap B, A\cap C, A\cap B\cap C\}\equiv \mathcal{P}(A)\cap (D^\Theta \setminus \emptyset)$, i.e. $D_1$ contains all the parts of $A$ which are included in $D^\Theta$, except the empty set. $D_2$ contains all elements which do not contain the letter $A$, i.e. $D_2 = (\{B,C\}, \cup, \cap) = D^{\{B,C\}} = \{B, C, B\cup C, B\cap C\}$. $D_3 = \{A\cup B, A\cup C, A\cup B\cup C, A\cup (B\cap C)\}$, i.e. In this case, sets whose formulas contain both the letters $A$ and at least a letter from $\{B,C\}$ and are not included in $D_1$.

 \item If one supposes the truth is in $A\cap B$, then one has $D1 = \{A\cap B, A\cap B\cap C\}$, $D_2=\{C\}$; i.e. $D_2$ elements do not contain the letters $A$, $B$; and $D_3 = \{A, B, A\cup B, A\cap C, B\cap C, \ldots \}$, i.e. what’s left from $D^\Theta \setminus \{ \emptyset\}$ after removing $D_1$ and $D_2$.

 \item If one supposes the truth is in $A\cup B$, then one has $D_1 = \{A, B, A\cap B, A\cup B\}$, and all other sets included in these four ones, i.e. $A\cap C$, $B\cap C$, $A\cap B\cap C$, $A\cup (B\cap C)$,
$B\cup (A\cap C)$, $(A\cap C)\cup (B\cap C)$, etc; $D_2 = \{C\}$,  i.e. $D_2$ elements do not contain the letters $A$, $B$ and $D_3 = \{A\cup C, B\cup C, A\cup B\cup C, C\cup (A\cap B)\}$.

 \item If one supposes the truth is in $A\cup B \cup C$, then one has $D_1 = D^\Theta \setminus \{\emptyset\}$. $D_2$ does not exist since $s(A\cup B\cup C) = \{A,B,C\}$ and $\Theta \setminus \{A,B,C\} = \emptyset$; i.e. $D_2$ elements do not contain the letters $A$, $B$, $C$. $D_3$ does not exist since $(D^\Theta \setminus \{\emptyset\}) \setminus D_1 = \emptyset$.

 \item If one supposes the truth is in $A\cap B \cap C$, then one has $D_1 = \{A\cap B\cap C\}$; $D_2$ does not exist; i.e. $D_2$ elements do not contain the letters $A$, $B$, $C$ and $D_3$ equals everything else, i.e. $(D^\Theta \setminus \{\emptyset\}) \setminus D_1 =
\{A, B, C, A\cap B, A\cap C, B\cap C, A\cup B, A\cup C, B\cup C, A\cup B\cup C,
A\cup (B\cap C), \ldots \}$; $D_3$ has $19-1-1 =17$ elements.

\end{itemize}

We propose several Belief Conditioning Rules (BCR)\index{BCR (Belief Conditioning Rule)} for deriving a (posterior) conditioning belief assignment $m(.|A)$ from a (prior) bba $m(.)$ and a conditioning set $A\in D^\Theta\setminus\{\emptyset\}$. For all forthcoming BCR formulas, of course we have:

\begin{equation}
m(X|A)=0, \qquad \text{if}\quad X\notin D_1
\end{equation}

\subsection{Belief Conditioning Rule no. 1 (BCR1)}
%-------------------------------------------------------------------

The Belief Conditioning Rule no. 1\index{BCR1 rule (Belief Conditioning Rule no. 1)}, denoted BCR1 for short, is defined for $X\in D_1$ by the formula 

\begin{equation}
\displaystyle
m_{BCR1}(X|A)=m(X) + \frac{m(X)\cdot\sum_{Z\in D_2\cup D_3} m(Z)}{\sum_{Y\in D_1}m(Y)}=\frac{m(X)}{\sum_{Y\in D_1}m(Y)}
\label{eqBCR1}
\end{equation}

This is the easiest transfer of masses of the elements from $D_2$ and $D_3$ to the non-empty elements from $D_1$. This transfer is done indiscriminately in a similar way to Dempster's rule\index{Dempster's rule} transfer, but this transfer is less exact. Therefore the sum of masses of non-empty elements from $D_2$ and $D_3$ is transferred to the masses of non-empty elements from $D_1$ proportionally with respect to their corresponding non-null masses.\\

% BCRs completionj
%--------------------------
In a set of sets, such as $D_1$, $D_2$, $D_3$, $D^\Theta$, we consider the {\it{inclusion}} of sets, $\subseteq$, which is a partial ordering relationship. The model of $D^\Theta$ generates submodels for $D_1$, $D_2$ and $D_3$ respectively.

Let $W\in D_3$. We say $X\in D_1$ is the {\it{$k$-largest}}, $k\geq 1$, element from $D_1$ that is included in $W$, if: $\nexists Y\in D_1\setminus \{X\}$  with $X\subset Y$, and $X\subset W$. Depending on the model, there are $k\geq 1$ such elements. Similarly, we say that $X\in D_1$ is the {\it{$k$-smallest}}, $k\geq 1$, element from $D_1$ that is included in $W$, if: $\nexists Y\in D_1\setminus \{X\}$ with $Y\subset X$, and $X\subset W$. Since in many cases there are $k\geq 1$ such elements, we call each of them a {\it{$k$-smallest element}}.

We recall that the {\it{DSm Cardinal}}, i.e. $Card_{DSm}(X)$ for
$X \in D^\Theta$, is the number of distinct parts that
compose $X$ in the Venn Diagram.  It depends on the
model and on the cardinal of $\Theta$, see  \cite{DSmTBook_2004a} for details.

\noindent
We partially increasingly order the elements in $D_1$ using the inclusion relationship and their DSm Cardinals. Since there are elements $X,Y \in D_1$ that are in no relationship with each other (i.e. $X\nsubseteq Y$, $Y\nsubseteq X$), but having the {\it{same}} DSm Cardinal, we list them together in a same class. We, similarly as in statistics, say that $X$ is a {\it{$k$-median}}, $k\geq 1$, element if $X$ is in the middle of $D_1$ in the case when cardinal of $D_1$, $Card(D_1)$, is odd, or if $Card(D_1)$ is even we take the left and right classes from the middle of $D_1$ list. We also compute a $k$-average, $k\geq 1$, element of $D_1$ by first computing $\sum_{X\in D_1}Card_{DSm}(X)/Card(D_1)$. Then {\it{$k$-average elements}} are those whose DSm Cardinal is close to the atomic average of $D_1$. For each computation of $k$-largest, $k$-smallest, $k$-median, or $k$-average we take the whole class of a such element. In a class as stated above, all elements have the same DSm Cardinal and none is included in another one.\\

\noindent
Let's see a few examples:
\begin{itemize}
\item[a)] Let $\Theta=\{A,B,C\}$, Shafer's model\index{Shafer's model}, and the truth is in $B\cup C$.
$$D_1=\{B,C,B\cup C\} \qquad Card_{DSm}(B)=Card_{DSm}(C)=1 \qquad Card_{DSm}(B\cup C)=2$$
\noindent
In $D_1$, we have: the $1$-largest element is $B\cup C$; the $k$-smallest (herein $2$-smallest) are $B$, $C$; the $k$-median (herein $2$-median) is the class of $C$, which is formed by the elements $B$, $C$; the $k$-average of $D_1$ is $(Card_{DSm}(B)+Card_{DSm}(C)+Card_{DSm}(B\cup C))/Card(D_1)=(1+1+2)/3= 1.333333 \approx 1$ and the $k$-averages are $B$, $C$.

\item[b)] Let $\Theta=\{A,B,C\}$, free DSm model\index{Dezert-Smarandache free model}, and the truth is in $B\cup C$. Then:
\begin{multline*}
D_1=\{
\underbrace{B\cap C\cap A}_{Card_{DSm}=1},
\underbrace{B\cap C, B\cap A, C\cap A}_{Card_{DSm}=2},\\
\underbrace{(B\cap C)\cup (B\cap A), (B\cap C)\cup (C\cap A), (B\cap A)\cup (C\cap A)}_{Card_{DSm}=3},\\
\underbrace{(B\cap C)\cup (B\cap A)\cup (C\cap A), B, C}_{Card_{DSm}=4},
\underbrace{B\cup (C\cap A), C\cup (B\cap A)}_{Card_{DSm}=5}, \underbrace{B\cup C}_{Card_{DSm}=6}
\}
 \end{multline*}
\noindent
Therefore $Card(D_1)=13$.

\noindent
$$D_2=\{ \underbrace{A}_{Card_{DSm}=4}\}\qquad \text{and} \qquad Card(D_2)=1.$$
\noindent
$$D_3=\{\underbrace{A\cup(B\cap C)}_{Card_{DSm}=5}, \underbrace{A\cup B, A\cup C}_{Card_{DSm}=6}, \underbrace{A\cup B \cup C}_{Card_{DSm}=7}\}\qquad \text{and} \qquad Card(D_3)=4.$$
\noindent
One verifies easily that:
$$Card(D^\Theta)=19=Card(D_1)+Card(D_2)+Card(D_3)+ 1 \text{ element (the empty set)}$$

\item[c)] Let $\Theta=\{A,B,C\}$, free DSm model\index{Dezert-Smarandache free model}, and the truth is in $B$.

$$D_1=\{\underbrace{B\cap C\cap A}_{\text{class 1}},\underbrace{B\cap A, B\cap C}_{\text{class 2}}, \underbrace{B}_{\text{class 3}}\} $$

$$Card_{DSm}(\text{class 1})=1 \qquad Card_{DSm}(\text{class 2})=2 \qquad Card_{DSm}(\text{class 3})=3$$

\noindent
The $D_1$ list is increasingly by class DSm Cardinals. The $1$-largest element is $B$; the $1$-smallest is $B\cap C\cap A$; the $2$-median elements are $B\cap A$, $B\cap C$; the average of DSm Cardinals is $[1\cdot(1) + 2\cdot (2) + 1\cdot (3)]/4=2$. The $2$-average elements are  $B\cap A$, $B\cap C$.

\end{itemize}

For the following BCR formulas, the $k$-largest, $k$-smallest, $k$-median, and $k$-average elements respectively are calculated {\it{only}} for those elements from $D_1$ that are included in a given $W$ (where $W\in D_3$), not for the whole $D_1$.

\subsection{Belief Conditioning Rule no. 2 (BCR2)}
%-------------------------------------------------------------------

In Belief Conditioning Rule no. 2\index{BCR2 rule (Belief Conditioning Rule no. 2)}, i.e. BCR2 for short, a better transfer is proposed. While the sum of masses of elements from $D_2$ is redistributed in a similar way to the non-empty elements from $D_1$ proportionally with respect to their corresponding non-null masses, the masses of elements from $D_3$ are redistributed differently, i.e. if $W\in D_3$ then its whole mass, $m(W)$, is transferred to the $k$-largest (with respect to inclusion from $D_1$) set $X\subset W$; this is considered a {\it{pessimistic/prudent way}}. The formula of BCR2\index{BCR2 rule (Belief Conditioning Rule no. 2)} for $X\in D_1$ is defined by:

\begin{equation}
m_{BCR2}(X|A)=m(X) + \frac{m(X)\cdot\sum_{Z\in D_2} m(Z)}{\sum_{Y\in D_1}m(Y)} + \sum_{
\begin{array}{c}
\scriptstyle W\in D_3\\
\scriptstyle X\subset W, X \,\text{is $k$-largest}
\end{array}}
m(W)/k
\end{equation}

\noindent
or equivalently

\begin{equation}
m_{BCR2}(X|A)=\frac{m(X)\cdot\sum_{Z\in D_1\cup D_2} m(Z)}{\sum_{Y\in D_1}m(Y)} + \sum_{
\begin{array}{c}
\scriptstyle W\in D_3\\
\scriptstyle X\subset W, X \,\text{is $k$-largest}
\end{array}} m(W)/k
\end{equation}

\noindent
where $X$ is the $k$-largest (with respect to inclusion) set included in $W$. The previous formula is actually equivalent in the Shafer's model to the following formula:

\begin{equation}
m_{BCR2}(X|A) =\frac{m(X)\cdot\sum_{Z\in D_1\cup D_2} m(Z)}{\sum_{Y\in D_1}m(Y)} + \sum_{
\begin{array}{c}
\scriptstyle W\in D_3\\
\scriptstyle W=X\,\text{when}\, \Theta\setminus s(A)\equiv\emptyset
\end{array}}
m(W)/k
\end{equation}

\subsection{Belief Conditioning Rule no. 3 (BCR3)}
%-------------------------------------------------------------------

The Belief Conditioning Rule no. 3\index{BCR3 rule (Belief Conditioning Rule no. 3)}, i.e. BCR3 is a dual of BCR2\index{BCR2 rule (Belief Conditioning Rule no. 2)}, but the transfer of $m(W)$ is done to the $k$-smallest, $k\geq 1$, (with respect to inclusion) set $X\subset W$, i.e. in an {\it{optimistic way}}. The formula of BCR3\index{BCR3 rule (Belief Conditioning Rule no. 3)} for $X\in D_1$ is defined by:

\begin{equation}
m_{BCR3}(X|A)=m(X) + \frac{m(X)\cdot\sum_{Z\in D_2} m(Z)}{\sum_{Y\in D_1}m(Y)} + \sum_{
\begin{array}{c}
\scriptstyle W\in D_3\\
\scriptstyle X\subset W, X \, \text{is $k$-smallest}
\end{array}}
m(W)/k
\end{equation}

\noindent
or equivalently

\begin{equation}
m_{BCR3}(X|A)=\frac{m(X)\cdot\sum_{Z\in D_1\cup D_2} m(Z)}{\sum_{Y\in D_1}m(Y)} + \sum_{
\begin{array}{c}
\scriptstyle W\in D_3\\
\scriptstyle X\subset W, X \, \text{is $k$-smallest}
\end{array}} m(W)/k
\end{equation}

\noindent
where $X$ is the $k$-smallest, $k\geq 1$, (with respect to inclusion) set included in $W$. \\

There are cases where BCR2\index{BCR2 rule (Belief Conditioning Rule no. 2)} and BCR3\index{BCR3 rule (Belief Conditioning Rule no. 3)} coincide, i.e. when there is only one, or none, $X\subset W$ for each $W\in D_3$.

\subsection{Belief Conditioning Rule no. 4 (BCR4)}
%-------------------------------------------------------------------

In an average between pessimistic and optimistic ways, we can consider "$X$ $k$-median" in the previous formulas in order to get the Belief Conditioning Rule no. 4 (BCR4)\index{BCR4 rule (Belief Conditioning Rule no. 4)}, i.e. for any $X\in D_1$,

\begin{equation}
m_{BCR4}(X|A)=m(X)+
\frac{m(X)\cdot\sum_{Z\in D_2} m(Z)}{\sum_{Y\in D_1}m(Y)}
+
\sum_{
\begin{array}{c}
\scriptstyle W\in D_3\\
\scriptstyle X\subset W, X \, \text{is $k$-median}
\end{array}} m(W)/k
\end{equation}

\noindent
or equivalently
\begin{equation}
m_{BCR4}(X|A)=\frac{m(X)\cdot\sum_{Z\in D1\cup D_2} m(Z)}{\sum_{Y\in D_1}m(Y)}
+
\sum_{
\begin{array}{c}
\scriptstyle W\in D_3\\
\scriptstyle X\subset W, X \, \text{is $k$-median}
\end{array}} m(W)/k
\end{equation}

\noindent
where $X$ is a $k$-median element of all elements from $D_1$ which are included in $W$. Here we do a medium (neither pessimistic nor optimistic) transfer.

\subsection{Belief Conditioning Rule no. 5 (BCR5)}
%-------------------------------------------------------------------

We replace "$X$ is $k$-median" by "$X$ is $k$-average" in BCR4\index{BCR4 rule (Belief Conditioning Rule no. 4)} formula in order to obtain the BCR5\index{BCR5 rule (Belief Conditioning Rule no. 5)}, i.e. for any $X\in D_1$,
\begin{equation}
m_{BCR5}(X|A)=\frac{m(X)\cdot\sum_{Z\in D1\cup D_2} m(Z)}{\sum_{Y\in D_1}m(Y)}
+
\sum_{
\begin{array}{c}
\scriptstyle W\in D_3\\
\scriptstyle X\subset W, X \, \text{is $k$-average}
\end{array}} m(W)/k
\end{equation}
\noindent
where $X$ is a $k$-average element of all elements from $D_1$ which are included in $W$. This transfer from $D_3$ is also medium and the result close to BCR4's\index{BCR4 rule (Belief Conditioning Rule no. 4)}.

\subsection{Belief Conditioning Rule no. 6 (BCR6)}
%-------------------------------------------------------------------

BCR6\index{BCR6 rule (Belief Conditioning Rule no. 6)} does a uniform redistribution of masses of each element $W\in D_3$ to all elements from $D_1$ which are included in $W$, i.e. for any $X\in D_1$,
\begin{equation}
m_{BCR6}(X|A)=\frac{m(X)\cdot\sum_{Z\in D1\cup D_2} m(Z)}{\sum_{Y\in D_1}m(Y)}
+
\sum_{
\begin{array}{c}
\scriptstyle W\in D_3\\
\scriptstyle X\subset W
\end{array}} 
\frac{m(W)}{Card\{V\in D_1 | V\subset W\}}
\end{equation}
\noindent
where $Card\{V\in D_1 | V\subset W\}$ is the cardinal (number) of $D_1$ sets included in $W$.

\subsection{Belief Conditioning Rule no. 7 (BCR7)} 
%-------------------------------------------------------------------

In our opinion, a better (prudent) transfer is done in the following Belief Conditioning Rule no. 7 (BCR7)\index{BCR7 rule (Belief Conditioning Rule no. 7)} defined for any $X\in D_1$ by:

\begin{multline*}
m_{BCR7}(X|A)=m(X)+ \displaystyle \frac{m(X)\cdot\sum_{Z\in D_2} m(Z)}{\sum_{Y\in D_1}m(Y)} \\
+ 
m(X)\cdot 
\displaystyle\sum_{
\begin{array}{c}
\scriptstyle W\in D_3\\
\scriptstyle X\subset W, S(W)\neq 0\\
\end{array}}
\frac{m(W)}{S(W)}
\quad +
\sum_{
\begin{array}{c}
\scriptstyle W\in D_3\\
\scriptstyle X\subset W, X \,\text{is $k$-largest}\\
\scriptstyle S(W)=0
\end{array}} m(W)/k
\end{multline*}

\noindent
where $S(W)\triangleq \displaystyle \sum_{\begin{array}{c}
\scriptstyle Y\in D_1, Y\subset W
\end{array}} m(Y)$.\\

\noindent
Or, simplified we get:
\begin{multline}
m_{BCR7}(X|A)=m(X)\cdot\Bigg[
 \displaystyle \frac{\sum_{Z\in D_1\cup D_2} m(Z)}{\sum_{Y\in D_1}m(Y)} 
+ 
 \displaystyle\sum_{
\begin{array}{c}
\scriptstyle W\in D_3\\
\scriptstyle X\subset W, \, S(W)\neq 0\\
\end{array}}
\frac{m(W)}{S(W)} 
\Bigg]\\
+
\sum_{
\begin{array}{c}
\scriptstyle W\in D_3\\
\scriptstyle X\subset W, X \,\text{is $k$-largest}\\
\scriptstyle S(W)=0
\end{array}} m(W)/k
\label{eq:BCR7}
\end{multline}

The transfer is done in the following way:
\begin{itemize}
\item the sum of masses of elements in $D_2$ are redistributed to the non-empty elements from $D_1$ proportionally with respect to their corresponding non-null masses (similarly as in BCR1\index{BCR1 rule (Belief Conditioning Rule no. 1)}-BCR6\index{BCR6 rule (Belief Conditioning Rule no. 6)} and BCR8-BCR11 defined in the sequel);
\item for each element $W\in D_3$, its mass $m(W)$ is distributed to all elements from $D_1$ which are included in $W$ and whose masses are non-null proportionally with their corresponding masses (according to the second term of the formula \eqref{eq:BCR7});
\item but, if all elements from $D_1$ which are included in $W$ have null masses, then $m(W)$ is transferred to the $k$-largest $X$ from $D_1$, which is included in $W$ (according to the last term of the formula \eqref{eq:BCR7}); this is the {\it{pessimistic/prudent way}}.
\end{itemize}

\subsection{Belief Conditioning Rule no. 8 (BCR8)} 
%-------------------------------------------------------------------

A dual of BCR7\index{BCR7 rule (Belief Conditioning Rule no. 7)} is the Belief Conditioning Rule no. 8 (BCR8)\index{BCR8 rule (Belief Conditioning Rule no. 8)}, where we consider the {\it{optimistic/more specialized way}}, i.e. "$X$ is $k$-largest" is replaced by "$X$ is $k$-smallest", $k\geq 1$ in \eqref{eq:BCR7}. Therefore, BCR8\index{BCR8 rule (Belief Conditioning Rule no. 8)} formula for any $X\in D_1$ is given by :

\begin{multline}
m_{BCR8}(X|A)=m(X)\cdot\Bigg[
 \displaystyle \frac{\sum_{Z\in D_1\cup D_2} m(Z)}{\sum_{Y\in D_1}m(Y)} 
+ 
 \displaystyle\sum_{
\begin{array}{c}
\scriptstyle W\in D_3\\
\scriptstyle X\subset W, \, S(W)\neq 0\\
\end{array}}
\frac{m(W)}{S(W)} 
\Bigg]\\
+
\sum_{
\begin{array}{c}
\scriptstyle W\in D_3\\
\scriptstyle X\subset W, X \,\text{is $k$-smallest}\\
\scriptstyle S(W)=0
\end{array}} m(W)/k
\label{eq:BCR8}
\end{multline}
\noindent
where $S(W)\triangleq \displaystyle \sum_{\begin{array}{c}
\scriptstyle Y\in D_1, Y\subset W
\end{array}} m(Y)$.

\subsection{Belief Conditioning Rule no. 9 (BCR9)} 
%-------------------------------------------------------------------

In an average between pessimistic and optimistic ways, we can consider "$X$ $k$-median" in the previous formulas \eqref{eq:BCR7} and \eqref{eq:BCR8} instead of "$k$-largest" or "$k$-smallest" in order to get the Belief Conditioning Rule no. 9 (BCR9)\index{BCR9 rule (Belief Conditioning Rule no. 9)}.

\subsection{Belief Conditioning Rule no. 10 (BCR10)}
%-------------------------------------------------------------------

BCR10\index{BCR10 rule (Belief Conditioning Rule no. 10)} is similar to BCR9\index{BCR9 rule (Belief Conditioning Rule no. 9)} using an average transfer (neither pessimistic nor optimistic) from $D_3$ to $D_1$. We only replace "$X$ $k$-median" by "$X$ $k$-average" in BCR9\index{BCR9 rule (Belief Conditioning Rule no. 9)} formula.

\subsection{Belief Conditioning Rule no. 11 (BCR11)}
%-------------------------------------------------------------------

BCR11\index{BCR11 rule (Belief Conditioning Rule no. 11)} does a uniform redistribution of masses of $D_3$ to $D_1$, as BCR6\index{BCR6 rule (Belief Conditioning Rule no. 6)}, but when $S(W)=0$ for $W\in D_3$. BCR11\index{BCR11 rule (Belief Conditioning Rule no. 11)} formula for any $X\in D_1$ is given by:

\begin{multline}
m_{BCR11}(X|A)=m(X)\cdot\Bigg[
 \displaystyle \frac{\sum_{Z\in D_1\cup D_2} m(Z)}{\sum_{Y\in D_1}m(Y)} 
+ 
 \displaystyle\sum_{
\begin{array}{c}
\scriptstyle W\in D_3\\
\scriptstyle X\subset W, \, S(W)\neq 0\\
\end{array}}
\frac{m(W)}{S(W)} 
\Bigg]\\
+
\sum_{
\begin{array}{c}
\scriptstyle W\in D_3\\
\scriptstyle X\subset W,\\
\scriptstyle S(W)=0
\end{array}} 
\frac{m(W)}{Card\{V\in D_1 | V\subset W\}}
\label{eq:BCR11}
\end{multline}
where $Card\{V\in D_1 | V\subset W\}$ is the cardinal (number) of $D_1$ sets included in $W$.

\subsection{More Belief Conditioning Rules (BCR12-BCR21)}
%--------------------------------------------------------------------------------

More versions of BCRs can be constructed that are distinguished through the way the masses of elements from $D_2\cup D_3$ are redistributed to those in $D_1$. So far, in BCR1\index{BCR1 rule (Belief Conditioning Rule no. 1)}-11\index{BCR11 rule (Belief Conditioning Rule no. 11)}, we have redistributed the masses of $D_2$ indiscriminately to $D_1$, but for the free and some hybrid DSm models\index{Dezert-Smarandache hybrid model} of $D^\Theta$ we can do a more exact redistribution.

There are elements in $D_2$ that don't include any element from $D_1$; the mass of these elements will be redistributed identically as in BCR1\index{BCR1 rule (Belief Conditioning Rule no. 1)}-\index{BCR11 rule (Belief Conditioning Rule no. 11)}. But other elements from $D_2$ that include at least one element from $D_1$ will be redistributed as we did before with $D_3$. So we can improve the last ten BCRs for any $X\in D_1$ as follows:

\begin{multline}
m_{BCR12}(X|A)=m(X) +
\bigl[ m(X)
\cdot \displaystyle\sum_{
\begin{array}{c}
\scriptstyle Z\in D_2\\
\scriptstyle \nexists Y\in D_1\, \text{with}\, Y\subset Z
\end{array}
} 
m(Z)\bigr] / \sum_{Y\in D_1}m(Y)\\
+  \displaystyle\sum_{
\begin{array}{c}
\scriptstyle Z\in D_2\\
\scriptstyle X\subset Z, \, $X$ \,\text{is $k$-largest}
\end{array}}
m(Z)/k 
+  \displaystyle\sum_{
\begin{array}{c}
\scriptstyle W\in D_3\\
\scriptstyle X\subset W, \, $X$ \,\text{is $k$-largest}
\end{array}}
m(W)/k 
\label{eq:BCR12}
\end{multline}

\noindent
or equivalently

\begin{multline}
m_{BCR12}(X|A)=
\bigl[ m(X)
\cdot \displaystyle\sum_{
\begin{array}{c}
\scriptstyle Z\in D_1,\\
\scriptstyle \text{or}\, Z\in D_2 \,\mid\, \nexists Y\in D_1\, \text{with}\, Y\subset Z
\end{array}
} 
m(Z)\bigr] / \sum_{Y\in D_1}m(Y)\\
+  \displaystyle\sum_{
\begin{array}{c}
\scriptstyle W\in D_2\cup D_3\\
\scriptstyle X\subset W, \, $X$ \,\text{is $k$-largest}
\end{array}}
m(W)/k 
\label{eq:BCR12bis}
\end{multline}

\begin{multline}
m_{BCR13}(X|A)=
\bigl[ m(X)
\cdot \displaystyle\sum_{
\begin{array}{c}
\scriptstyle Z\in D_1,\\
\scriptstyle \text{or}\, Z\in D_2 \,\mid\, \nexists Y\in D_1\, \text{with}\, Y\subset Z
\end{array}
} 
m(Z)\bigr] / \sum_{Y\in D_1}m(Y)\\
+  \displaystyle\sum_{
\begin{array}{c}
\scriptstyle W\in D_2\cup D_3\\
\scriptstyle X\subset W, \, $X$ \,\text{is $k$-smallest}
\end{array}}
m(W)/k 
\label{eq:BCR13}
\end{multline}

\begin{multline}
m_{BCR14}(X|A)=
\bigl[ m(X)
\cdot \displaystyle\sum_{
\begin{array}{c}
\scriptstyle Z\in D_1,\\
\scriptstyle \text{or}\, Z\in D_2 \,\mid\, \nexists Y\in D_1\, \text{with}\, Y\subset Z
\end{array}
} 
m(Z)\bigr] / \sum_{Y\in D_1}m(Y)\\
+  \displaystyle\sum_{
\begin{array}{c}
\scriptstyle W\in D_2\cup D_3\\
\scriptstyle X\subset W, \, $X$ \,\text{is $k$-median}
\end{array}}
m(W)/k 
\label{eq:BCR14}
\end{multline}

\begin{multline}
m_{BCR15}(X|A)=
\bigl[ m(X)
\cdot \displaystyle\sum_{
\begin{array}{c}
\scriptstyle Z\in D_1,\\
\scriptstyle \text{or}\, Z\in D_2 \,\mid\, \nexists Y\in D_1\, \text{with}\, Y\subset Z
\end{array}
} 
m(Z)\bigr] / \sum_{Y\in D_1}m(Y)\\
+  \displaystyle\sum_{
\begin{array}{c}
\scriptstyle W\in D_2\cup D_3\\
\scriptstyle X\subset W, \, $X$ \,\text{is $k$-average}
\end{array}}
m(W)/k 
\label{eq:BCR15}
\end{multline}

\begin{multline}
m_{BCR16}(X|A)=
\bigl[ m(X)
\cdot \displaystyle\sum_{
\begin{array}{c}
\scriptstyle Z\in D_1,\\
\scriptstyle \text{or}\, Z\in D_2 \,\mid\, \nexists Y\in D_1\, \text{with}\, Y\subset Z
\end{array}
} 
m(Z)\bigr] / \sum_{Y\in D_1}m(Y)\\
+  \displaystyle\sum_{
\begin{array}{c}
\scriptstyle W\in D_2\cup D_3\\
\scriptstyle X\subset W
\end{array}}
\frac{m(W) }{Card\{V\in D_1 | V\subset W\}}
\label{eq:BCR16}
\end{multline}

\begin{multline}
m_{BCR17}(X|A)=
m(X)\cdot \Bigg[ \bigl[  
\displaystyle\sum_{
\begin{array}{c}
\scriptstyle Z\in D_1,\\
\scriptstyle \text{or}\, Z\in D_2 \,\mid\, \nexists Y\in D_1\, \text{with}\, Y\subset Z
\end{array}
} 
m(Z)\bigr] / \sum_{Y\in D_1}m(Y)
+ 
\displaystyle\sum_{
\begin{array}{c}
\scriptstyle W\in D_2 \cup D_3\\
\scriptstyle X\subset W\\
\scriptstyle S(W)\neq 0
\end{array}
} 
\frac{m(W)}{S(W)}
\Bigg]\\
+  \displaystyle\sum_{
\begin{array}{c}
\scriptstyle W\in D_2\cup D_3\\
\scriptstyle X\subset W, \, $X$ \,\text{is $k$-largest}\\
\scriptstyle S(W)=0
\end{array}}
m(W)/k
\label{eq:BCR17}
\end{multline}

\begin{multline}
m_{BCR18}(X|A)=
m(X)\cdot \Bigg[ \bigl[  
\displaystyle\sum_{
\begin{array}{c}
\scriptstyle Z\in D_1,\\
\scriptstyle \text{or}\, Z\in D_2 \,\mid\, \nexists Y\in D_1\, \text{with}\, Y\subset Z
\end{array}
} 
m(Z)\bigr] / \sum_{Y\in D_1}m(Y)
+ 
\displaystyle\sum_{
\begin{array}{c}
\scriptstyle W\in D_2 \cup D_3\\
\scriptstyle X\subset W\\
\scriptstyle S(W)\neq 0
\end{array}
} 
\frac{m(W)}{S(W)}
\Bigg]\\
+  \displaystyle\sum_{
\begin{array}{c}
\scriptstyle W\in D_2\cup D_3\\
\scriptstyle X\subset W, \, $X$ \,\text{is $k$-smallest}\\
\scriptstyle S(W)=0
\end{array}}
m(W)/k
\label{eq:BCR18}
\end{multline}

\begin{multline}
m_{BCR19}(X|A)=
m(X)\cdot \Bigg[ \bigl[  
\displaystyle\sum_{
\begin{array}{c}
\scriptstyle Z\in D_1,\\
\scriptstyle \text{or}\, Z\in D_2 \,\mid\, \nexists Y\in D_1\, \text{with}\, Y\subset Z
\end{array}
} 
m(Z)\bigr] / \sum_{Y\in D_1}m(Y)
+ 
\displaystyle\sum_{
\begin{array}{c}
\scriptstyle W\in D_2 \cup D_3\\
\scriptstyle X\subset W\\
\scriptstyle S(W)\neq 0
\end{array}
} 
\frac{m(W)}{S(W)}
\Bigg]\\
+  \displaystyle\sum_{
\begin{array}{c}
\scriptstyle W\in D_2\cup D_3\\
\scriptstyle X\subset W, \, $X$ \,\text{is $k$-median}\\
\scriptstyle S(W)=0
\end{array}}
m(W)/k
\label{eq:BCR19}
\end{multline}

\begin{multline}
m_{BCR20}(X|A)=
m(X)\cdot \Bigg[ \bigl[  
\displaystyle\sum_{
\begin{array}{c}
\scriptstyle Z\in D_1,\\
\scriptstyle \text{or}\, Z\in D_2 \,\mid\, \nexists Y\in D_1\, \text{with}\, Y\subset Z
\end{array}
} 
m(Z)\bigr] / \sum_{Y\in D_1}m(Y)
+ 
\displaystyle\sum_{
\begin{array}{c}
\scriptstyle W\in D_2 \cup D_3\\
\scriptstyle X\subset W\\
\scriptstyle S(W)\neq 0
\end{array}
} 
\frac{m(W)}{S(W)}
\Bigg]\\
+  \displaystyle\sum_{
\begin{array}{c}
\scriptstyle W\in D_2\cup D_3\\
\scriptstyle X\subset W, \, $X$ \,\text{is $k$-average}\\
\scriptstyle S(W)=0
\end{array}}
m(W)/k
\label{eq:BCR20}
\end{multline}

\begin{multline}
m_{BCR21}(X|A)=
m(X)\cdot \Bigg[ \bigl[  
\displaystyle\sum_{
\begin{array}{c}
\scriptstyle Z\in D_1,\\
\scriptstyle \text{or}\, Z\in D_2 \,\mid\, \nexists Y\in D_1\, \text{with}\, Y\subset Z
\end{array}
} 
m(Z)\bigr] / \sum_{Y\in D_1}m(Y)
+ 
\displaystyle\sum_{
\begin{array}{c}
\scriptstyle W\in D_2 \cup D_3\\
\scriptstyle X\subset W\\
\scriptstyle S(W)\neq 0
\end{array}
} 
\frac{m(W)}{S(W)}
\Bigg]\\
+  \displaystyle\sum_{
\begin{array}{c}
\scriptstyle W\in D_2\cup D_3\\
\scriptstyle X\subset W\\
\scriptstyle S(W)=0
\end{array}}
\frac{m(W) }{Card\{V\in D_1 | V\subset W\}}
\label{eq:BCR21}
\end{multline}

Surely, other combinations of the ways of redistributions of masses from $D_2$ and $D_3$ to $D_1$ can be done, obtaining new BCR rules.

% ============================================================================
\section{Examples}
% ============================================================================

\subsection{Example no. 1 (free DSm model with non-Bayesian bba)}

Let's consider $\Theta=\{A,B,C\}$, the free DSm model\index{Dezert-Smarandache free model} (no intersection is empty) and the following prior bba 

$$m_1(A)=0.2\qquad m_1(B)=0.1\qquad m_1(C)=0.2 \qquad m_1(A\cup B)=0.1\qquad m_1(B\cup C)=0.1$$
$$m_1(A\cup(B\cap C))=0.1\qquad m_1(A\cap B)=0.1 \qquad m_1(A\cup B\cup C)=0.1$$
\noindent
and let's assume that the truth is in $B\cup C$, i.e. the conditioning term is $B\cup C$. Then:
\begin{multline*}
D_1=\{
\underbrace{B\cap C\cap A}_{Card_{DSm}=1},
\underbrace{B\cap C, B\cap A, C\cap A}_{Card_{DSm}=2},\\
\underbrace{(B\cap C)\cup (B\cap A), (B\cap C)\cup (C\cap A), (B\cap A)\cup (C\cap A)}_{Card_{DSm}=3},\\
\underbrace{(B\cap C)\cup (B\cap A)\cup (C\cap A), B, C}_{Card_{DSm}=4},
\underbrace{B\cup (C\cap A), C\cup (B\cap A)}_{Card_{DSm}=5}, \underbrace{B\cup C}_{Card_{DSm}=6}
\}
 \end{multline*}

\noindent
Therefore $Card(D_1)=13$.\\

We recall that $\forall X\in D^\Theta$, the {\it{DSm Cardinal}} of $X$, $Card_{DSm}(X)$, is equal to the number of distinct parts that compose $X$ in the Venn Diagram (see below) according to the given model on $D^\Theta$. By definition, $Card_{DSm}(\emptyset)=0$ (see \cite{DSmTBook_2004a} for examples and details).
\begin{figure}[h]
\centering
{\tt \setlength{\unitlength}{1pt}
\begin{picture}(90,90)
\thinlines    
\put(40,60){\circle{40}}
\put(60,60){\circle{40}}
\put(50,40){\circle{40}}
\put(15,84){\vector(1,-1){10}}
\put(7,84){$A$}
\put(84,84){\vector(-1,-1){10}}
\put(85,84){$B$}
\put(74,15){\vector(-1,1){10}}
\put(75,10){$C$}
\end{picture}}
\caption{Venn Diagram for the 3D free DSm model }
 \label{BCR:FreeDSmModel}
 \end{figure}
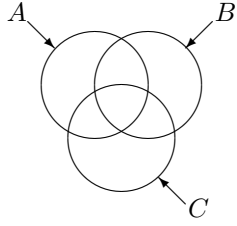

\noindent
$$D_2=\{ \underbrace{A}_{Card_{DSm}=4}\}\qquad \text{and} \qquad Card(D_2)=1.$$

\noindent
$$D_3=\{\underbrace{A\cup(B\cap C)}_{Card_{DSm}=5}, \underbrace{A\cup B, A\cup C}_{Card_{DSm}=6}, \underbrace{A\cup B \cup C}_{Card_{DSm}=7}\}\qquad \text{and} \qquad Card(D_3)=4.$$

\noindent
One verifies easily that:
$$Card(D^\Theta)=19=Card(D_1)+Card(D_2)+Card(D_3)+ 1 \text{ element (the empty set)}$$

\noindent
The masses of elements from $D_2\cup D_3$ are transferred to the elements of $D_1$. The ways these transfers are done make the distinction between the BCRs.

\begin{enumerate}
\item[a)] In BCR1, the sum of masses of $D_2$ and $D_3$ are indiscriminately distributed to $B$, $C$, $B\cup C$, $A\cap B$, proportionally to their corresponding masses 0.1, 0.2, 0.1, and respectively 0.1, i.e.

$$m(D_2\cup D_3)=m_1(A)+m_1(A\cup B) + m_1(A\cup (B\cap C)) +  m_1(A\cup B\cup C)=0.5$$
$$\frac{x_B}{0.1}=\frac{y_C}{0.2}=\frac{z_{B\cup C}}{0.1}=\frac{w_{B\cap A}}{0.1}=\frac{0.5}{0.5}=1$$
\noindent
whence $x_B=0.1$, $y_C=0.2$, $z_{B\cup C}=0.1$ and $w_{B\cap A}=0.1$ are added to the original masses of $B$, $C$, $B\cup C$ and $B\cap A$ respectively.

\noindent
Finally, one gets with BCR1-based conditioning:
\begin{align*}
& m_{BCR1}(B|B\cup C)=0.2\\
& m_{BCR1}(C|B\cup C)=0.4\\
& m_{BCR1}(B\cup C|B\cup C)=0.2\\
& m_{BCR1}(B\cap A|B\cup C)=0.2
\end{align*}

\item[b)] In BCR2, $m(D_2)=m_1(A)=0.2$ and is indiscriminately distributed to $B$, $C$, $B\cup C$, $A\cap B$, proportionally to their corresponding masses, i.e.
$$\frac{x_B}{0.1}=\frac{y_C}{0.2}=\frac{z_{B\cup C}}{0.1}=\frac{w_{B\cap A}}{0.1}=\frac{0.2}{0.5}=0.4$$
\noindent
whence $x_B=0.04$, $y_C=0.08$, $z_{B\cup C}=0.04$ and $w_{B\cap A}=0.04$.

$m(D_3)$ is redistributed, element by element, to the $k$-largest $D_1$ element in each case: 
$m_1(A\cup B)=0.1$ to $B\cup (C\cap A)$, since $B\cup (C\cap A)\in D_1$ and it is the 1-largest one from $D_1$ included in $A\cup B$; 
$m_1(A\cup (B\cap C))=0.1$ to $(B\cap A) \cup (C\cap A)\cup (B\cap C)$ for a similar reason; $m_1(A\cup B\cup C)=0.1$ to $B\cup C$.
\noindent
Finally, one gets with BCR2-based conditioning:
\begin{align*}
& m_{BCR2}(B|B\cup C)=0.14\\
& m_{BCR2}(C|B\cup C)=0.28\\
& m_{BCR2}(B\cup C|B\cup C)=0.24\\
& m_{BCR2}(B\cap A|B\cup C)=0.14\\
& m_{BCR2}((B\cap A)\cup (C\cap A)\cup (B\cap C)|B\cup C)=0.10\\
& m_{BCR2}(B\cup (C\cap A)|B\cup C)=0.10
\end{align*}

\item[c)] In BCR3, instead of $k$-largest $D_1$ elements, we consider $k$-smallest ones.
$m(D_2)=m_1(A)=0.2$ is exactly distributed as in BCR2. But $m(D_3)$ is, in each case, redistributed to the $k$-smallest $D_1$ element, which is $A\cap B\cap C$ (1-smallest herein). Hence:
\begin{align*}
& m_{BCR3}(B|B\cup C)=0.14\\
& m_{BCR3}(C|B\cup C)=0.28\\
& m_{BCR3}(B\cup C|B\cup C)=0.14\\
& m_{BCR3}(B\cap A|B\cup C)=0.14\\
& m_{BCR3}(A\cap B\cap C|B\cup C)=0.30
\end{align*}

\item[d)] In BCR4, we use $k$-median.
\begin{itemize}
\item $A\cup B$ includes the following $D_1$ elements:
\begin{multline*}
A\cap B \cap C, B \cap C, B \cap A, C \cap A, (B \cap C)\cup (B \cap A), 
\underbrace{\mid}_{\text{\tiny{median herein}}}\\
  (B \cap C)\cup (C \cap A), (B \cap A)\cup (C \cap A), (B \cap C)\cup (B \cap A)\cup (C\cap A), B, B\cup (A\cap C)
\end{multline*}

Hence we take the whole class: $(B \cap C)\cup (B \cap A)$, $(B \cap C)\cup (C \cap A)$, $ (B \cap A)\cup (C \cap A)$, i.e. 3-medians; each one receiving 1/3 of $0.1=m_1(A\cup B)$.

\item $A\cup (B\cap C)$ includes the following $D_1$ elements:
\begin{multline*}
A\cap B \cap C, B \cap C, B \cap A, C \cap A,  \underbrace{\mid}_{\text{\tiny{median herein}}} \\
(B \cap C)\cup (B \cap A), (B \cap C)\cup (C \cap A), (B \cap A)\cup (C \cap A), (B \cap C)\cup (B \cap A)\cup (C\cap A)
\end{multline*}
\noindent
Hence we take the left and right (to the median) classes: $B \cap C$, $B \cap A$, $C \cap A$, $(B \cap C)\cup (B \cap A)$, $(B \cap C)\cup (C \cap A)$, $(B \cap A)\cup (C \cap A)$, i.e. 6-medians, each ones receiving 1/6 of $0.1=m_1(A\cup (B\cap C))$.

\item $A\cup B\cup C$ includes all $D_1$ elements, hence the 3-medians are $(B \cap C)\cup (B \cap A)$, $(B \cap C)\cup (C \cap A)$ and $(B \cap A)\cup (C \cap A)$; each one receiving 1/3 of $0.1=m_1(A\cup B\cup C)$
\end{itemize}

Totalizing, one finally gets:
\begin{align*}
& m_{BCR4}(B|B\cup C)=42/300\\
& m_{BCR4}(C|B\cup C)=84/300\\
& m_{BCR4}(B\cup C|B\cup C)=42/300\\
& m_{BCR4}(B\cap A|B\cup C)=47/300\\
& m_{BCR4}(B\cap C|B\cup C)=5/300 \\
& m_{BCR4}(C\cap A|B\cup C)=5/300\\
& m_{BCR4}((B\cap C)\cup (B\cap A)|B\cup C)=25/300 \\
&  m_{BCR4}((B\cap C)\cup (C\cap A)|B\cup C)=25/300\\
& m_{BCR4}((B\cap A)\cup (C\cap A)|B\cup C)=25/300
\end{align*}

\item[e)] In BCR5, we compute $k$-average, i.e. the $k$-average of DSm cardinals of the $D_1$ elements included in eack $W\in D_3$.
\begin{itemize}
\item For $A\cup B$, using the results got in BCR4 above:
$$\sum_{X\in D_1, X\subset A\cup B} Card_{DSm}(X)= 1+ 3\cdot (2)+3\cdot (3)+4+4+5=29$$
\noindent
The average DSm cardinal per element is $29/10=2.9\approx 3$. Hence $(B \cap C)\cup (B \cap A)$, $(B \cap C)\cup (C \cap A)$, $(B \cap A)\cup (C \cap A)$, i.e. the 3-average elements, receive each 1/3 of $0.1=m_1(A\cup B)$.
\item For $A\cup (B\cap C)$, one has
$$\sum_{X\in D_1, X\subset A\cup (B\cap C)} Card_{DSm}(X)= 1+ 3\cdot (2)+3\cdot (3)+4=20$$
\noindent
The average DSm cardinal per element is $20/8=2.5\approx 3$. Hence again $(B \cap C)\cup (B \cap A)$, $(B \cap C)\cup (C \cap A)$, $(B \cap A)\cup (C \cap A)$, i.e. the 3-average elements, receive each 1/3 of $0.1=m_1(A\cup (B\cap C))$.
\item For $A\cup B\cup C$, one has
$$\sum_{X\in D_1, X\subset A\cup B\cup C} Card_{DSm}(X)= 1+ 3\cdot (2)+3\cdot (3)+4+2\cdot (4)+2\cdot (5)+6=44$$
\noindent
The average DSm cardinal per element is $44/13=3.38\approx 3$. Hence $(B \cap C)\cup (B \cap A)$, $(B \cap C)\cup (C \cap A)$, $(B \cap A)\cup (C \cap A)$, i.e. the 3-average elements, receive each 1/3 of $0.1=m_1(A\cup B\cup C)$.
\end{itemize}

Totalizing, one finally gets:
\begin{align*}
& m_{BCR5}(B|B\cup C)=42/300\\
& m_{BCR5}(C|B\cup C)=84/300\\
& m_{BCR5}(B\cup C|B\cup C)=42/300\\
& m_{BCR5}(B\cap A|B\cup C)=42/300\\
& m_{BCR5}((B\cap C)\cup (B\cap A)|B\cup C)=30/300\\
& m_{BCR5}((B\cap C)\cup (C\cap A)|B\cup C)=30/300\\
& m_{BCR5}((B\cap A)\cup (C\cap A)|B\cup C)=30/300
\end{align*}

\item[f)] In BCR6, the $k$-average is replaced by uniform redistribution of $D_3$ elements' masses to all $D_1$ elements included in each $W\in D_3$.

\begin{itemize}
\item The mass $m_1(A\cup B)=0.1$ is equally split among each $D_1$ element included in $A\cup B$ (see the list of them in BCR4 above), hence 1/10 of 0.1 to each.
\item Similarly, $m_1(A\cup (B\cap C))=0.1$ is equally split among each $D_1$ element included in $A\cup (B\cap C)$, hence 1/8 of 0.1 to each.
\item And, in the same way, $m_1(A\cup B\cup C)=0.1$ is equally split among each $D_1$ element included in $A\cup B\cup C$, hence 1/13 of 0.1 to each.
\end{itemize}

Totalizing, one finally gets:
\begin{align*}
& m_{BCR6}(B|B\cup C)=820/5200\\
& m_{BCR6}(C|B\cup C)=1996/5200\\
& m_{BCR6}(B\cup C|B\cup C)=768/5200\\
& m_{BCR6}(B\cap A|B\cup C)=885/5200\\
& m_{BCR6}(A\cap B\cap C|B\cup C)=157/5200\\
& m_{BCR6}(B\cap C|B\cup C)=157/5200 \\
& m_{BCR6}(C\cap A|B\cup C)=157/5200\\
& m_{BCR6}((B\cap C)\cup (B\cap A)|B\cup C)=157/5200\\
& m_{BCR6}((B\cap C)\cup (C\cap A)|B\cup C)=157/5200\\
& m_{BCR6}((B\cap A)\cup (C\cap A)|B\cup C)=157/5200\\
& m_{BCR6}((B\cap C)\cup (B\cap A)\cup (C\cap A) |B\cup C)=157/5200 \\
& m_{BCR6}(B\cup (C\cap A)|B\cup C)=92/5200\\
& m_{BCR6}(C\cup (B\cap A)|B\cup C)=40/5200
\end{align*}

\item[g)] In BCR7, $m(D_2)$ is also indiscriminately redistributed, but $m(D_3)$ is redistributed in a different more refined way.
\begin{itemize}
\item The mass $m_1(A\cup B)=0.1$ is transferred to $B$ and $B\cap A$ since these are the only $D_1$ elements included in  $A\cup B$ whose masses are non-zero, proportionally to their corresponding masses, i.e.
$$\frac{x_B}{0.1}=\frac{w_{B\cap A}}{0.1}=\frac{0.1}{0.2}=0.5$$
\noindent
whence $x_B=0.05$ and $w_{B\cap A}=0.05$.
\item
$m_1(A\cup (B\cap C))=0.1$ is transferred to $B\cap A$ only since no other $D_1$ element with non-zero mass is included in $A\cup (B\cap C)$.
\item 
$m_1(A\cup B\cup C)=0.1$ is similarly transferred to $B$, $C$, $B\cap A$, $B\cup C$, i.e.
$$\frac{x_B}{0.1}=\frac{y_C}{0.2}=\frac{z_{B\cup C}}{0.1}=\frac{w_{B\cap A}}{0.1}=\frac{0.1}{0.5}=0.2$$
\noindent
whence $x_B=0.02$, $y_C=0.04$, $z_{B\cup C}=0.02$ and $w_{B\cap A}=0.02$.
\end{itemize}

Totalizing, one finally gets:
\begin{align*}
& m_{BCR7}(B|B\cup C)=0.21\\
& m_{BCR7}(C|B\cup C)=0.32\\
& m_{BCR7}(B\cup C|B\cup C)=0.16\\
& m_{BCR7}(B\cap A|B\cup C)=0.31
\end{align*}

\item[h)] In BCR8-11, since there is no $W\in D_3$ such that the sum of masses of $D_1$ elements included in $W$ be zero, i.e. $s(W)\neq 0$, we can not deal with "$k$-elements", hence the results are identical to BCR7.

\item[i)] In BCR12, $m(D_2)$ is redistributed differently. $m_1(A)=0.2$ is transferred to $(A\cap B)\cup (A\cap C)$ since this is the 1-largest $D_1$ element included in $A$. $m(D_3)$ is transferred exactly as in BCR2. Finally, one gets:
\begin{align*}
& m_{BCR12}(B|B\cup C)=0.1\\
& m_{BCR12}(C|B\cup C)=0.2\\
& m_{BCR12}(B\cup C|B\cup C)=0.2\\
& m_{BCR12}(B\cap A|B\cup C)=0.1\\
& m_{BCR12}((B\cap A)\cup (C\cap A)\cup (B\cap C)|B\cup C)=0.1\\
& m_{BCR12}((A\cap B)\cup (A\cap C) |B\cup C)=0.1 \\
& m_{BCR12}(B\cup (C\cap A)|B\cup C)=0.1
\end{align*}

\item[j)] In BCR13, $m(D_2)$ is redistributed to the 1-smallest, i.e. to $A\cap B\cap C$ and $m(D_3)$ is redistributed as in BCR3. Therefore one gets:
\begin{align*}
& m_{BCR13}(B|B\cup C)=0.1\\
& m_{BCR13}(C|B\cup C)=0.2\\
& m_{BCR13}(B\cup C|B\cup C)=0.1\\
& m_{BCR13}(B\cap A|B\cup C)=0.1\\
& m_{BCR13}(A\cap B\cap C|B\cup C)=0.5
\end{align*}

\item[k)] In BCR14, $m_1(A)=0.2$, where $A\in D_2$, is redistributed to the $k$-medians of $A\cap B\cap C$, $B\cap A$, $C\cap A$, $(B\cap A)\cup (C\cap A)$ which are included in $A$ and belong to $D_1$. The 2-medians are $B\cap A$, $C\cap A$, hence each receives 1/2 of 0.2. $m(D_3)$ is redistributed as in BCR4. Therefore one gets:
\begin{align*}
& m_{BCR14}(B|B\cup C)=30/300\\
& m_{BCR14}(C|B\cup C)=60/300\\
& m_{BCR14}(B\cup C|B\cup C)=30/300\\
& m_{BCR14}(B\cap A|B\cup C)=65/300\\
& m_{BCR14}(B\cap C|B\cup C)=5/300\\
& m_{BCR14}(C\cap A|B\cup C)=35/300\\
& m_{BCR14}((B\cap C)\cup (B\cap A)|B\cup C)=25/300\\
& m_{BCR14}((B\cap C)\cup (C\cap A)|B\cup C)=25/300\\
& m_{BCR14}((B\cap A)\cup (C\cap A)|B\cup C)=25/300
\end{align*}

\item[l)] In BCR15, $m_1(A)=0.2$, where $A\in D_2$, is redistributed to the $k$-averages.
\begin{multline*}
\frac{1}{4}\cdot [Card_{DSm}(A\cap B\cap C)+Card_{DSm}(B\cap C)+ Card_{DSm}(C\cap A) + Card_{DSm}((B\cap A)\cup(C\cap A))]\\
=\frac{1+2+2+3}{4}=2
\end{multline*}
\noindent
Hence each of $B\cap C$, $C\cap A$ receives 1/2 of 2; $m(D_3)$ is redistributed as in BCR5. Therefore one gets:
\begin{align*}
& m_{BCR15}(B|B\cup C)=3/30\\
& m_{BCR15}(C|B\cup C)=6/30\\
& m_{BCR15}(B\cup C|B\cup C)=3/30\\
& m_{BCR15}(B\cap A|B\cup C)=6/30\\
& m_{BCR15}(C\cap A|B\cup C)=3/30\\
& m_{BCR15}((B\cap C)\cup (B\cap A)|B\cup C)=3/30\\
& m_{BCR15}((B\cap C)\cup (C\cap A)|B\cup C)=3/30\\
& m_{BCR15}((B\cap A)\cup (C\cap A)|B\cup C)=3/30
\end{align*}

\item[m)] In BCR16, $m_1(A)=0.2$, where $A\in D_2$, is uniformly transferred to all $D_1$ elements included in $A$, i.e. to $A\cap B\cap C$, $B\cap A$, $C\cap A$, $(B\cap A)\cup (C\cap A)$, hence each one receives 1/4 of 0.2. $m(D_3)$ is redistributed as in BCR6. Therefore one gets:
\begin{align*}
& m_{BCR16}(B|B\cup C)=612/5200\\
& m_{BCR16}(C|B\cup C)=1080/5200\\
& m_{BCR16}(B\cup C|B\cup C)=560/5200\\
& m_{BCR16}(B\cap A|B\cup C)=937/5200\\
& m_{BCR16}(A\cap B\cap C|B\cup C)=417/5200\\
& m_{BCR16}(B\cap C|B\cup C)=157/5200 \\
& m_{BCR16}(C\cap A|B\cup C)=417/5200\\
& m_{BCR16}((B\cap C)\cup (B\cap A)|B\cup C)=157/5200\\
& m_{BCR16}((B\cap C)\cup (C\cap A)|B\cup C)=157/5200\\
& m_{BCR16}((B\cap A)\cup (C\cap A)|B\cup C)=417/5200\\
& m_{BCR16}((B\cap C)\cup (B\cap A)\cup (C\cap A) |B\cup C)=157/5200 \\
& m_{BCR16}(B\cup (C\cap A)|B\cup C)=92/5200\\
& m_{BCR16}(C\cup (B\cap A)|B\cup C)=40/5200
\end{align*}

\item[n)] In BCR17, $m_1(A)=0.2$, where $A\in D_2$, is transferred to $B\cap A$ since $B\cap A\subset A$ and $m_1(B\cap A)> 0$. No other $D_1$ element with non-zero mass is included in $A$. $m(D_3)$ is redistributed as in BCR7. Therefore one gets:
\begin{align*}
& m_{BCR17}(B|B\cup C)=0.17\\
& m_{BCR17}(C|B\cup C)=0.24\\
& m_{BCR17}(B\cup C|B\cup C)=0.12\\
& m_{BCR17}(B\cap A|B\cup C)=0.47
\end{align*}

\item[o)] BCR18-21 give the same result as BCR17 since no $k$-elements occur in these cases.

\item[p)] SCR does not work for free DSm models. But we can use the extended (from the power set $2^\Theta$ to the hyper-power set $D^\Theta$) Dempster's rule\index{extended Dempster's rule} (see Daniel's Chapter \cite{MilanChapter1-2006}) in order to combine $m_1(.)$ with $m_2(B\cup C)=1$, because the truth is in $B\cup C$, as in Shafer's conditioning rule. But since we have a free DSm model, no transfer is needed, hence Dempster's rule is reduced to DSm Classic rule (DSmC), which is a generalization of conjunctive rule. One gets:
\begin{align*}
& m_{DSmC}(B|B\cup C)=0.1\\
& m_{DSmC}(C|B\cup C)=0.2\\
& m_{DSmC}(B\cup C|B\cup C)=0.2\\
& m_{DSmC}(B\cap A|B\cup C)=0.1\\
& m_{DSmC}((A\cap B)\cup (A\cap C)|B\cup C)=0.2\\
& m_{DSmC}(B\cup (A\cap C)|B\cup C)=0.1\\
& m_{DSmC}((A\cap B)\cup (B\cap C)\cup (C\cap A)|B\cup C)=0.1
\end{align*}

In the free DSm model, if the truth is in $A$, BCR12 gives the same result as $m_1(.)$ fusioned with $m_2(A)=1$ using the classic DSm rule.

\end{enumerate}

\subsection{Example no. 2 (Shafer's model with non-Bayesian bba)} 

Let's consider $\Theta=\{A,B,C\}$ with Shafer's model\index{Shafer's model} and the following prior bba:
$$m_1(A)=0.2\qquad m_1(B)=0.1\qquad m_1(C)=0.2$$
$$m_1(A\cup B)=0.1 \qquad m_1(B\cup C)=0.1 \qquad m_1(A\cup B\cup C)=0.3$$

\noindent
Let's assume as conditioning constraint that the truth is in $B\cup C$. $D^\Theta$ is decomposed into
$$D_1=\{B,C,B\cup C\}$$
$$D_2=\{A\}$$
$$D_3=\{A\cup B, A\cup C, A\cup B\cup C\}$$

\noindent
The Venn Diagram corresponding to Shafer's model for this example is given in Figure \ref{BCR:ShaferModel} below.
\begin{figure}[h]
\centering
{\tt \setlength{\unitlength}{1pt}
\begin{picture}(90,90)
\thinlines    
\put(20,60){\circle{40}} 
\put(80,60){\circle{40}} 
\put(50,10){\circle{40}}
\put(-5,84){\vector(1,-1){10}}
\put(-13,84){$A$}
\put(104,84){\vector(-1,-1){10}}
\put(105,84){$B$}
\put(85,10){\vector(-1,0){15}}
\put(87,7){$C$}
\end{picture}}
\vspace{2mm}
\caption{Venn Diagram for the 3D Shafer's model }
 \label{BCR:ShaferModel}
 \end{figure}
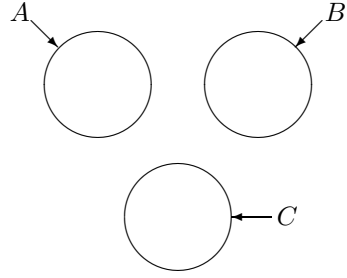

\begin{enumerate}

\item[a)] In BCR1, $m(D_2\cup D_3)=m_1(A)+m_1(A\cup B)+m_1(A\cup B\cup C)=0.6$ is redistributed to $B$, $C$, $B\cup C$, proportionally to their corresponding masses 0.1, 0.2, 0.1 respectively, i.e.
$$\frac{x_B}{0.1}=\frac{y_C}{0.2}=\frac{z_{B\cup C}}{0.1}=\frac{0.6}{0.4}=1.5$$
\noindent
whence $x_B=0.15$, $y_C=0.30$, $z_{B\cup C}=0.15$ are added to the original masses of $B$, $C$, $B\cup C$ respectively. Finally, one gets with BCR1-based conditioning:
\begin{align*}
& m_{BCR1}(B|B\cup C)=0.25\\
& m_{BCR1}(C|B\cup C)=0.50\\
& m_{BCR1}(B\cup C|B\cup C)=0.25
\end{align*}

\item[b)] In BCR2, $m(D_2)=m_1(A)=0.2$ and is indiscriminately distributed to $B$, $C$ and $B\cup C$ proportionally to their corresponding masses, i.e.
$$\frac{x_B}{0.1}=\frac{y_C}{0.2}=\frac{z_{B\cup C}}{0.1}=\frac{0.2}{0.4}=0.5$$
\noindent
whence $x_B=0.05$, $y_C=0.10$, and $z_{B\cup C}=0.05$.

For $D_3$, $m_1(A\cup B)=0.1$ is transferred to $B$ (1-largest), $m_1(A\cup B\cup C)=0.3$ is transferred to $A\cup B$. Finally, one gets with BCR2-based conditioning:
\begin{align*}
& m_{BCR2}(B|B\cup C)=0.25\\
& m_{BCR2}(C|B\cup C)=0.30\\
& m_{BCR2}(B\cup C|B\cup C)=0.45
\end{align*}

\item[c)] In BCR3 for $D_3$, $m_1(A\cup B)=0.1$ is transferred to $B$ (1-smallest), $m_1(A\cup B\cup C)=0.3$ is transferred to $B$, $C$ (2-smallest). Finally, one gets with BCR3-based conditioning:
\begin{align*}
& m_{BCR3}(B|B\cup C)=0.40\\
& m_{BCR3}(C|B\cup C)=0.45\\
& m_{BCR3}(B\cup C|B\cup C)=0.15
\end{align*}

\item[d)] In BCR4 for $D_3$, $m_1(A\cup B)=0.1$ is transferred to $B$ (1-median), $m_1(A\cup B\cup C)=0.3$ is transferred to $B$, $C$ (2-medians). Finally, one gets same result as with BCR3, i.e.
\begin{align*}
& m_{BCR4}(B|B\cup C)=0.40\\
& m_{BCR4}(C|B\cup C)=0.45\\
& m_{BCR4}(B\cup C|B\cup C)=0.15
\end{align*}

\item[e)] In BCR5 for $D_3$, $m_1(A\cup B)=0.1$ is transferred to $B$. Let's compute
$$
\frac{1}{3}\cdot [Card_{DSm}(B)+Card_{DSm}(C)+ Card_{DSm}(B\cup C) ]
=\frac{1+1+2}{3}\approx 1
$$
\noindent
Hence 2-averages are $B$ and $C$. So with BCR5, one gets same result as with BCR3, i.e.
\begin{align*}
& m_{BCR5}(B|B\cup C)=0.40\\
& m_{BCR5}(C|B\cup C)=0.45\\
& m_{BCR5}(B\cup C|B\cup C)=0.15
\end{align*}

\item[f)] In BCR6 for $D_3$, $m_1(A\cup B)=0.1$ is transferred to $B$ (the only $D_1$ element included in $A\cup B$), $m_1(A\cup B\cup C)=0.3$ is transferred to $B$, $C$, $B\cup C$, each one receiving 1/3 of 0.3.  Finally, one gets 
\begin{align*}
& m_{BCR6}(B|B\cup C)=0.35\\
& m_{BCR6}(C|B\cup C)=0.40\\
& m_{BCR6}(B\cup C|B\cup C)=0.25
\end{align*}

\item[g)] In BCR7 for $D_3$, $m_1(A\cup B)=0.1$ is transferred to $B$ since $B\subset A\cup B$ and $m(B)> 0$; $m_1(A\cup B\cup C)=0.3$ is transferred to $B$, $C$, $B\cup C$ proportionally to their corresponding masses:
$$\frac{x_B}{0.1}=\frac{y_C}{0.2}=\frac{z_{B\cup C}}{0.1}=\frac{0.3}{0.4}=0.75$$
\noindent
whence $x_B=0.075$, $y_C=0.15$, and $z_{B\cup C}=0.075$. Finally, one gets 
\begin{align*}
& m_{BCR7}(B|B\cup C)=0.325\\
& m_{BCR7}(C|B\cup C)=0.450\\
& m_{BCR7}(B\cup C|B\cup C)=0.225
\end{align*}

\item[h)] BCR8-11 give the same result as BCR7 in this example, since there is no case of $k$-elements.

\item[i)] In BCR12: For $D_2$ for all BCR12-21, $m_1(A)=0.2$ is redistributed to $B$, $C$, $B\cup C$ as in BCR2. $m(D_3)$ is redistributed as in BCR2. The result is the same as in BCR2.

\item[j)] BCR13-15 give the same result  as in BCR3.

\item[k)] BCR16 gives the same result  as in BCR6.

\item[l)] BCR17-21: For $D_3$, $m_1(A\cup B)=0.1$ is transferred to $B$ (no case of $k$-elements herein); $m_1(A\cup B\cup C)=0.3$ is transferred to $B$, $C$, $B\cup C$ proportionally to their corresponding masses as in BCR7. Therefore one gets same result as in BCR7, i.e.
\begin{align*}
& m_{BCR17}(B|B\cup C)=0.325\\
& m_{BCR17}(C|B\cup C)=0.450\\
& m_{BCR17}(B\cup C|B\cup C)=0.225
\end{align*}

\item[m)] BCR22, 23, 24, 25, 26 give the same results as BCR7, 8, 9, 10, 11 respectively since $D_2$ is indiscriminately redistributed to $D_1$ elements.

\item[n)] BCR27, 28, 29, 30, 31 give the same results as BCR2, 3, 4, 5, 6 respectively for the same reason as previously.

\item[o)] If one applies the SCR, i.e. one combines with Dempster's rule $m_1(.)$ with $m_2(B\cup C)=1$, because the truth is in $B\cup C$ as Glenn Shafer proposes, one gets:
\begin{align*}
& m_{SCR}(B|B\cup C)=0.25\\
& m_{SCR}(C|B\cup C)=0.25\\
& m_{SCR}(B\cup C|B\cup C)=0.50
\end{align*}

\end{enumerate}

\subsection{Example no. 3 (Shafer's model with Bayesian bba)} 
%**********************************************************************************

Let's consider $\Theta=\{A,B,C,D\}$ with Shafer's model\index{Shafer's model} and the following prior Bayesian bba\index{Bayesian basic belief assignment}:
$$m_1(A)=0.4\qquad m_1(B)=0.1\qquad m_1(C)=0.2 \qquad m_1(D)=0.3$$

\noindent
Let's assume that one finds out that the truth is in $C\cup D$. From formulas of BCRs conditioning rules one gets the same result for all the BCRs in such example according to the following table
\begin{table}[h]
\centering
%\small
\begin{tabular}{|c|cccc|}
\hline
  &     $A$      & $B$       & $C$   & $D$   \\
\hline
$m_1(.)$ & 0.4 & 0.1 & 0.2 & 0.3\\
\hline 
$m_{BCR1-31}(.|C\cup D)$ & 0 & 0 & 0.40 & 0.60\\
 \hline
\end{tabular}
\caption{Conditioning results based on BCRs given the truth is in $C\cup D$.}
\label{Table3BCRs}
\end{table}

Let's examine the conditional bba obtained directly from the {\it{fusion}} of the prior bba $m_1(.)$ with the belief assignment focused only on $C\cup D$, say $m_2(C\cup D)=1$ using three main rules of combination (Dempster's rule\index{Dempster's rule}, DSmH\index{Dezert-Smarandache Hybrid rule (DSmH)} and PCR5\index{PCR5 rule (Prop. Conflict Redist.)}). After elementary derivations, one gets final results given in Table \ref{Table3bisBCRs}. In the Bayesian case, all BCRs and Shafer's conditioning rule\index{SCR (Shafer's Conditioning Rule)} (with Dempster's rule\index{Dempster's rule}) give the same result.

\begin{table}[!h]
\centering
\small
\begin{tabular}{|r|ccccccc|}
\hline
  &     $A$      & $B$       & $C$   & $D$     & $C\cup D$ & $A\cup C\cup D$ & $B\cup C\cup D$\\
\hline
$m_{DS}(.|C\cup D)$ & 0 & 0 & 0.40 & 0.60 & 0 & 0 & 0\\
$m_{DSmH}(.|C\cup D)$ & 0 & 0 & 0.20 & 0.30 & 0 & 0.40 & 0.10\\
$m_{PCR5}(.|C\cup D)$ & 0.114286 & 0.009091 & 0.20 & 0.30 & 0.376623 & 0 & 0\\
\hline 
\end{tabular}
\caption{Conditioning results based on Dempster's, DSmH and PCR5 fusion rules.}
\label{Table3bisBCRs}
\end{table}

\section{Classification of the BCRs}

Let's note:

\noindent
$D_2^u=$ Redistribution of the whole $D_2$ is done undifferentiated to $D_1$

\noindent
$D_3^u=$ Redistribution of the whole $D_3$ is done undifferentiated to $D_1$

\noindent
$D_2^p=$ Redistribution of $D_2$ is particularly done from each $Z\in D_2$ to specific elements in $D_1$

\noindent
$D_3^p=$ Redistribution of $D_3$ is particularly done from each $W\in D_3$ to specific elements in $D_1$

\noindent
$D_2^s=$ $D_2$ is split into two disjoint subsets: one whose elements have the property that $s(W)\neq 0$, an another one such that its elements have $s(W)=0$. Each subset is differently redistributed to $D_1$

\noindent
$D_3^s=$ $D_3$ is similarly split into two disjoint subsets, that are redistributed as in $D_2^s$.\\

\noindent
Thus, we can organize and classify the BCRs as in Table \ref{TableBCRsClassif}. 
\begin{table}[h]
\centering
\small
\begin{tabular}{|ccc|}
\hline
 Ways of redistribution &     Belief Conditioning Rule & Specific Elements\\
\hline
$D_2^u$,$D_3^u$ & $BCR1$ & 
\\
\hline 
$D_2^u, D_3^p$ & 
$\begin{cases}
\fbox{BCR2}\\
BCR3\\
BCR4\\
BCR5\\
BCR6
\end{cases}$
 & 
$\begin{cases}
k-\text{largest}\\
k-\text{smallest}\\
k-\text{median}\\
k-\text{average}\\
\text{uniform distribution}
 \end{cases}$
\\
 \hline
$D_2^u, D_3^s$ & $\begin{cases}
\fbox{BCR7}\\
BCR8\\
BCR9\\
BCR10\\
BCR11
\end{cases}$
 & 
$\begin{cases}
k-\text{largest}\\
k-\text{smallest}\\
k-\text{median}\\
k-\text{average}\\
\text{uniform distribution}
 \end{cases}$
\\
 \hline
$D_2^p, D_3^p$ & $\begin{cases}
\fbox{BCR12}\\
BCR13\\
BCR14\\
BCR15\\
BCR16
\end{cases}$
 & 
$\begin{cases}
k-\text{largest}\\
k-\text{smallest}\\
k-\text{median}\\
k-\text{average}\\
\text{uniform distribution}
 \end{cases}$
\\
 \hline
 $D_2^s, D_3^s$ & $\begin{cases}
\fbox{BCR17}\\
BCR18\\
BCR19\\
BCR20\\
BCR21
\end{cases}$
 & 
$\begin{cases}
k-\text{largest}\\
k-\text{smallest}\\
k-\text{median}\\
k-\text{average}\\
\text{uniform distribution}
 \end{cases}$
\\
 \hline
\end{tabular}
\caption{Classification of Belief Conditioning Rules}
\label{TableBCRsClassif}
\end{table}
%
%\noindent
Other belief conditioning rules could also be defined according to Table \ref{TableBCRsClassif2}. But in our opinions, the most detailed and exact transfer is done by BCR17. So, we suggest to use preferentially BCR17 for a pessimistic/prudent view on conditioning problem and a more refined redistribution of conflicting masses, or BCR12 for a very pessimistic/prudent view and less
refined redistribution. If the Shafer's models holds for the frame under consideration, BCR12-21 will coincide with BCR2-11.\\

\begin{table}[h]
\centering
\small
\begin{tabular}{|ccc|}
\hline
 Ways of redistribution &     Belief Conditioning Rule & Specific Elements\\
\hline
$D_2^p, D_3^s$ & 
$\begin{cases}
\fbox{BCR22}\\
BCR23\\
BCR24\\
BCR25\\
BCR26
\end{cases}$
 & 
$\begin{cases}
k-\text{largest}\\
k-\text{smallest}\\
k-\text{median}\\
k-\text{average}\\
\text{uniform distribution}
 \end{cases}$
\\
 \hline
 $D_2^s, D_3^p$ & $\begin{cases}
\fbox{BCR27}\\
BCR28\\
BCR29\\
BCR30\\
BCR31
\end{cases}$
 & 
$\begin{cases}
k-\text{largest}\\
k-\text{smallest}\\
k-\text{median}\\
k-\text{average}\\
\text{uniform distribution}
 \end{cases}$
\\
 \hline
\end{tabular}
\caption{More Belief Conditioning Rules}
\label{TableBCRsClassif2}
\end{table}

In summary, the best among these BCR1-31, that we recommend to use, are: BCR17 for a pessimistic/prudent view on conditioning problem and a more refined redistribution of conflicting masses, or BCR12 for a very pessimistic/prudent view and less refined redistribution.\\

\noindent
BCR17 does the most refined redistribution of all BCR1-31, i.e. 
\newline - the mass $m(W)$ of each element $W$ in $D_2\cup D_3$ is transferred to those $X\in D_1$ elements which are
included in $W$ if any proportionally with respect to their non-empty masses;
\newline - if no such $X$ exists, the mass $m(W)$ is transferred in a pessimistic/prudent way to the $k$-largest elements from $D_1$ which are included in $W$ (in equal parts) if any;
\newline - if neither this way is possible, then $m(W)$ is indiscriminately distributed to all $X \in D_1$
proportionally with respect to their nonzero masses.\\

\noindent
BCR12 does the most pessimistic/prudent redistribution of all BCR1-31, i.e.:
\newline - the mass $m(W)$ of each $W$ in $D_2\cup D_3$ is transferred in
a pessimistic/prudent way to the $k$-largest elements $X$
from $D_1$ which are included in $W$ (in equal parts) if any;
\newline -	if this way is not possible, then $m(W)$ is
indiscriminately distributed to all $X$ from $D_1$ proportionally with respect their nonzero masses.\\

BCR12 is simpler than BCR17. BCR12 can be regarded as a generalization of SCR from the power set to the hyper-power set in the free DSm free model (all intersections non-empty).  In this case the result of BCR12 is equal to that of $m_1(.)$ combined with $m_2(A)=1$, when the truth is in $A$, using the DSm Classic fusion rule.

% ========================
\section{Properties for all BCRs}
% ========================

\label{secBCRProp}

\begin{enumerate}

\item For any $X\notin \mathcal{P}_{\mathcal{D}}(A)=D_1$, one has $m_{BCR}(X|A)=0$ by definition.

\item One has:
$$\sum_{X\in\mathcal{P}_{\mathcal{D}}(A)} m_{BCR}(X|A)=1$$
This can be proven from the fact that $\sum_{X\in D^\Theta}m(X)=1$. and $D^\Theta\setminus\{\emptyset\}=D_1\cup D_2\cup D_3$, where $D_1$, $D_2$ and $D_3$ have no element in common two by two. Since all masses of all elements from $D_2$ and $D_3$ are transferred to the non-empty elements of $D_1$using BCRs, no mass is lost neither gained, hence the total sum of masses remains equal to 1.
\item  Let $\Theta=\{\theta_1,\theta_2,\ldots, \theta_n\}$ and $A= \theta_1\cup \theta_2 \cup \ldots\cup \theta_n$ be the total ignorance.
Then, $m_{BCR1-31}(X|A) = m(X)$ for all $X$ in $D^\Theta$, because $D^\Theta \setminus \{\emptyset\}$ coincides with $D_1$. Hence there is no mass to be transferred from $D_2$ or $D_3$
to $D_1$ since $D_2$ and $D_3$ do not exist (are empty).

\item 
This property reduces all BRCs to the Bayesian formula: $m_{BCR}(X|A) = m(X\cap A)/m(A)$ for the trivial Bayesian case when focal elements are only singletons (no unions, neither intersections) and the truth is in one singleton only.

\noindent{\it{Proof}}: Let's consider $\Theta = \{\theta_1, \theta_2, ...,\theta_n\}$, $n\geq 2$, and all $\theta_{i}$ not empty, and $D^\Theta \equiv \Theta$. Let's have a bba $m(.): D^\Theta \mapsto [0,1]$. Without loss of generality, suppose the truth is in $\theta_1$ where $m(\theta_1)>0$. Then $m_{BCR}(\theta_1|\theta_1)=1$ and $m_{BCR}(X|\theta_1)=0$ for all $X$ different from $\theta_1$.
Then $1=m_{BCR}(\theta_1|\theta_1)=m(\theta_1\cap \theta_1)/ m(\theta_1)=1$, and for $i\neq 1$, we have $0=m_{BCR}(\theta_i|\theta_1)=m(\emptyset)/m(\theta_1)=0$.

\item 
In the Shafer's model, and a Bayesian bba $m(.)$, all
BCR1-31 coincide with SCR. In this case the conditioning with BCRs and fusioning with Dempster's rule commute.

\noindent
{\it{Proof}}: In a general case we can prove it as follows:  Let $\Theta = \{\theta_1, \theta_2, \ldots, \theta_{n}\}$, $n\geq 2$, and without loss of generality let’s suppose the truth is in $T = \theta_1\cup \theta_2\cup \ldots \cup \theta_{p}$, for $1\leq p\leq n$. Let’s consider two Bayesian masses $m_1(.)$ and $m_2(.)$. Then we can consider all other elements $\theta_{p+1}$, \ldots, $\theta_{n}$ as empty sets and in consequence the sum of their masses as the mass of the empty set (as in Smets’ open world).  BCRs\index{BCR1 rule (Belief Conditioning Rule no. 1)} work now exactly as (or we can say it is reduced to) Dempster's rule\index{Dempster's rule} redistributing this empty set mass to the elements in $T$ proportionally with their nonzero corresponding mass.  $D_1 = \{\theta_1, \theta_2, \ldots, \theta_{p}\}$,
$D_2 = \{\theta_{p+1}, \ldots,  \theta_{n}\}$, $D_3$ does not exist. And redistributing in $m_1(.|T)$ this empty sets' mass to non-empty sets $\theta_1$, $\theta_2$, \ldots , $\theta_{p}$ using BCRs\index{BCR1 rule (Belief Conditioning Rule no. 1)} is equivalent to combining $m_1(.)$ with $m_S(\theta_1\cup \theta_2\cup \ldots\cup \theta_{p})=1$. Similarly for $m_2(.|T)$. Since Dempter's fusion rule and Shafer's conditioning rule\index{SCR (Shafer's Conditioning Rule)} commute and BCRs\index{BCR1 rule (Belief Conditioning Rule no. 1)} are reduced to Dempster's rule\index{Dempster's rule} in a Shafer's model\index{Shafer's model} and Bayesian case, then BCRs\index{BCR1 rule (Belief Conditioning Rule no. 1)} commute with Dempster’s fusion rule in this case. QED

\item 
In the free DSm model, BCR12 can be regarded as a generalization of SCR from the power set to the
hyper-power set.  The result of BCR12 conditioning of a mass $m_1(.)$, when the truth is in $A$, is equal to that of fusioning $m_1(.)$ with $m_2(A)=1$, using the DSm Classic Rule.

\end{enumerate}

% ================
\section{Open question on conditioning versus fusion}
% ================

It is not to difficult too verify that fusion rules and conditioning rules do not commute in general, except in Dempster-Shafer Theory because Shafer's fusion and conditioning rules are based on the same operator\footnote{Proof of commutation between the Shafer’s conditioning rule and Dempster’s rule:
Let $m_1(.)$ be a bba and $m_S(A) = 1$.  Then, because Dempster’s rule, denoted $\oplus$, is associative we have $(m_1\oplus m_S)\oplus (m_2\oplus m_S) = m_1\oplus (m_S\oplus m_2)\oplus m_S$ and because it is commutative we get $m_1\oplus (m_2\oplus m_S)\oplus m_S$ and again because it is associative we have: $(m_1\oplus m_2)\oplus (m_S\oplus m_S)$;  hence, since $m_S\oplus m_S = m_S$, it is equal to: $(m_1\oplus m_2)\oplus m_S = m_1\oplus m_2\oplus m_S$, QED.} (Dempster's rule\index{Dempster's rule}), which make derivation very simple and appealing. 

We however think that things may be much more complex in reality than what has been proposed up to now if we follow our interpretation of belief conditioning and do not  see the belief conditioning as just a simple fusion of the prior bba with a bba focused on the conditioning event where the truth is (subjectively) supposed to be. From our belief conditioning interpretation, we make a strong difference between the fusion of several sources of evidences  (i.e. combination of bba's) and the conditioning of a given belief assignment according some extra knowledge (carrying some objective/absolute truth on a given subset) on the model itself. In our opinion, the conditioning must be interpreted as a revision of bba according to new integrity constraint on the truth of the space of the solutions. Based on this new idea on conditioning, we are face to a new and very important open question which can be stated as follows\footnote{The question can be extended for more than two sources actually.}: 

Let's consider two prior bba's $m_1(.)$ and $m_2(.)$ provided by two (cognitively) independent sources of evidences defined on $D^\Theta$ for a given model $\mathcal{M}$ (free, hybrid or Shafer's model\index{Shafer's model}) and then let's assume that the truth is known to be later on in a subset $A\in D^\Theta$, how to compute the combined conditional belief?\\

There are basically two possible answers to this question depending on the order the fusion and the conditioning are carried out. Let's denote by $\oplus$ the generic symbol for fusion operator (PCR5\index{PCR5 rule (Prop. Conflict Redist.)}, DSmH\index{Dezert-Smarandache Hybrid rule (DSmH)} or whatever) and by $Cond(.)$ the generic symbol for conditioning operator (typically BCRs).

\begin{enumerate}
\item Answer 1 (Fusion followed by conditioning (FC)):  
\begin{equation}
m_{FC}(.|A)= Cond(m_1(.) \oplus m_2(.))
\end{equation}

\item Answer 2 (Conditioning followed by the fusion (CF)): 
\begin{equation}
m_{CF}(.|A)= \underbrace{Cond(m_1(.))}_{m_1(.|A)} \oplus \underbrace{Cond(m_2(.))}_{m_2(.|A)}
\end{equation}

\end{enumerate}

Since in general\footnote{Because none of the new fusion and conditioning rules developed up to now satisfies the commutativity, but Dempster's rule.} the conditioning and the fusion do not commute, $m_{FC}(.|A)\neq m_{CF}(.|A)$, the fundamental open question arises:  How to justify the choice for one answer with respect to the other one (or maybe with respect to some other answers if any) to compute the combined conditional bba from $m_1(.)$, $m_2(.)$ and any conditioning subset $A$?\\

The only argumentation (maybe) for justifying the choice of $m_{FC}(.|A)$ or $m_{CF}(.|A)$ is only imposed by the possible temporal/sequential processing of sources and extra knowledge one receives, i.e. if one gets first $m_1(.)$ and $m_2(.)$ and later one knows that the truth is in $A$ then $m_{FC}(.|A)$ seems intuitively suitable, but if one gets first $m_1(.)$ and $A$, and later $m_2(.)$, then $m_{CF}(.|A)$ looks in better agreement with the chronology of information one has received in that case. If we make abstraction of temporal processing, then this fundamental and very difficult question remains unfortunately totally open.

\subsection{Examples of non commutation of BCR with fusion}

\subsubsection{Example no. 1 (Shafer's model and Bayesian bba's)} 

Let's consider $\Theta=\{A,B,C\}$ with Shafer's model\index{Shafer's model} and the following prior Bayesian bba's\index{Bayesian basic belief assignment}

$$m_1(A)=0.2\qquad m_1(B)=0.6 \qquad m_1(C)=0.2$$
$$m_2(A)=0.1\qquad m_2(B)=0.4 \qquad m_2(C)=0.5$$

Let's suppose one finds out the truth is in $A\cup B$ and let's examine the results $m_{CF}(.|A\cup B)$ and $m_{FC}(.|A\cup B)$ obtained from either the conditioning followed by the fusion, or the fusion followed by the conditioning.

\begin{itemize}
\item Case 1 : BCRs-based Conditioning followed by the PCR5-based Fusion

Using BCRs for conditioning, the mass $m_1(C)=0.2$ is redistributed to $A$ and $B$ proportionally to the masses 0.2 and 0.6 respectively; thus $x/0.2=y/0.6=0.2/(0.2+0.6)=1/4$ and therefore $x=0.2\cdot (1/4)=0.05$ is added to $m_1(A)$, while $y=0.6\cdot (1/4)=0.15$ is added to $m_1(B)$. Hence, one finally gets

$$m_1(A|A\cup B)=0.25\qquad m_1(B|A\cup B)=0.75 \qquad m_1(C|A\cup B)=0$$

\noindent
Similarly, the conditioning of $m_2(.)$ using the BCRs, will provide
$$m_2(A|A\cup B)=0.2\qquad m_2(B|A\cup B)=0.8 \qquad m_2(C|A\cup B)=0$$

\noindent
If one combines $m_1(.|A\cup B)$ and $m_2(.|A\cup B)$ with PCR5\index{PCR5 rule (Prop. Conflict Redist.)} fusion rule, one gets\footnote{We specify explicitly in notations $m_{CF}(.)$ and $m_{FC}(.)$ the type of the conditioning and fusion rules used for convenience, i.e $m_{C_{\tiny{BCRs}}F_{\tiny{PCR5}}}(.)$ means that the conditioning is based on BCRs and the fusion is based on PCR5.} 
$$m_{C_{\tiny{BCRs}}F_{\tiny{PCR5}}}(A|A\cup B)=0.129198\quad m_{C_{\tiny{BCRs}}F_{\tiny{PCR5}}}(B|A\cup B)=0.870802$$
\end{itemize}

\begin{itemize}
\item Case 2 : PCR5-based Fusion followed by the BCRs-based Conditioning

If one combines first $m_1(.)$ and $m_2(.)$ with PCR5\index{PCR5 rule (Prop. Conflict Redist.)} fusion rule, one gets
$$m_{PCR5}(A)=0.090476\quad m_{PCR5}(B)=0.561731 \quad m_{PCR5}(C)=0.347793$$

\noindent
and if one applies any of BCR rules for conditioning the combined prior $m_{PCR5}(.)$, one finally gets
$$m_{F_{\tiny{PCR5}}C_{\tiny{BCRs}}}(A|A\cup B)=0.138723\qquad m_{F_{\tiny{PCR5}}C_{\tiny{BCRs}}}(B|A\cup B)=0.861277$$
% \qquad m_{FC}(C|A\cup B)=0$$
\end{itemize}

From cases 1 and 2, one has proved that there exists at least one example for which
PCR5 fusion and BCRs conditioning do not commute since 
$$m_{F_{\tiny{PCR5}}C_{\tiny{BCRs}}}(.|A\cup B)\neq m_{C_{\tiny{BCRs}}F_{\tiny{PCR5}}}(.|A\cup B).$$

\begin{itemize}
\item Case 3 : BCRs-based Conditioning followed by Dempster's rule-based Fusion

If we consider the same masses $m_1(.)$ and $m_2(.)$ and if we apply the BCRs to each of them, one gets

$$m_1(A|A\cup B)=0.25\qquad m_1(B|A\cup B)=0.75 \qquad m_1(C|A\cup B)=0$$
$$m_2(A|A\cup B)=0.20\qquad m_2(B|A\cup B)=0.80 \qquad m_2(C|A\cup B)=0$$

\noindent
then if one combines them with Dempster’s rule, one finally gets

$$m_{C_{\tiny{BCRs}}F_{\tiny{DS}}}(A|A\cup B)=0.076923\quad m_{C_{\tiny{BCRs}}F_{\tiny{DS}}}(B|A\cup B)=0.923077$$

\end{itemize}

\begin{itemize}
\item Case 4 : Dempster's rule based Fusion followed by BCRs-based Conditioning

If we apply first the fusion of $m_1(.)$ with $m_2(.)$ with Dempster's rule of combination, one gets
$$m_{DS}(A)=0.055555\quad m_{DS}(B)= 0.666667 \quad m_{DS}(C)=0.277778$$
and if one applies BCRs for conditioning the prior $m_{DS}(.)$, one finally gets
$$m_{F_{\tiny{DS}}C_{\tiny{BCRs}}}(A|A\cup B)=0.076923\qquad m_{F_{\tiny{DS}}C_{\tiny{BCRs}}}(B|A\cup B)= 0.923077$$

From cases 3 and 4, we see that all BCRs (i.e. BCR1-BCR31) commute with Dempster’s fusion rule in a Shafer’s model and Bayesian case since:

$$m_{F_{\tiny{DS}}C_{\tiny{BCRs}}}(.|A\cup B)=m_{C_{\tiny{BCRs}}F_{\tiny{DS}}}(.|A\cup B).$$

But this is a trivial result because in this specific case (Shafer's model with Bayesian bba's), we know (cf 
Property 5 in Section \ref{secBCRProp}) that BCRs coincide with SCR and already know that SCR commutes with Dempter's fusion rule.

\end{itemize}

\subsubsection{Example no. 2 (Shafer's model and non Bayesian bba's)} 
%**********************************************************************************

Let's consider $\Theta=\{A,B,C\}$ with Shafer's model\index{Shafer's model} and the following prior non Bayesian bba's\index{Bayesian basic belief assignment}
$$m_1(A)=0.3\qquad m_1(B)=0.1 \qquad m_1(C)=0.2\qquad m_1(A\cup B)=0.1 \qquad m_1(B\cup C)=0.3$$
$$m_2(A)=0.1\qquad m_2(B)=0.2 \qquad m_2(C)=0.3 \qquad m_2(A\cup B)=0.2 \qquad m_2(B\cup C)=0.2$$

Let's suppose one finds out the truth is in $B\cup C$ and let's examine the results $m_{CF}(.|B\cup C)$ and $m_{FC}(.|B\cup C)$ obtained from either the conditioning followed by the fusion, or the fusion followed by the conditioning. In this second example we only provide results for BCR12 and BCR17 since we consider them as the most appealing BCR rules. We decompose $D^\Theta$ into $D_1=\{B,C,B\cup C\}$, $D_2=\{A\}$ and $D_3=\{A\cup B\}$.

\begin{itemize}
\item Case 1 : BCR12/BCR17-based Conditioning followed by the PCR5-based Fusion

Using BCR12 or BCR17 for conditioning $m_1(.)$ and $m_2(.)$, one gets herein the same result with both BCRs for each conditional bba, i.e.
$$m_1(B|B\cup C)=0.25\qquad m_1(C|B\cup C)=0.30 \qquad m_1(B\cup C|B\cup C)=0.45$$
$$m_2(B|B\cup C)=15/35\qquad m_2(C|B\cup C)=12/35 \qquad m_2(B\cup C|B\cup C)=8/35$$
\noindent
If one combines $m_1(.|B\cup C)$ and $m_2(.|B\cup C)$ with PCR5\index{PCR5 rule (Prop. Conflict Redist.)} fusion rule, one gets 
$$m_{C_{\tiny{BCR17}}F_{\tiny{PCR5}}}(.|B\cup C)=m_{C_{\tiny{BCR12}}F_{\tiny{PCR5}}}(.|B\cup C)$$ \noindent with
\begin{align*}
& m_{C_{\tiny{BCR12}}F_{\tiny{PCR5}}}(B|B\cup C)=0.446229\\
& m_{C_{\tiny{BCR12}}F_{\tiny{PCR5}}}(C|B\cup C)=0.450914\\
& m_{C_{\tiny{BCR12}}F_{\tiny{PCR5}}}(B\cup C|B\cup C)=0.102857
\end{align*}
\end{itemize}

\begin{itemize}
\item Case 2 : PCR5-based Fusion followed by BCR12/BCR17-based Conditioning

If one combines first $m_1(.)$ and $m_2(.)$ with PCR5\index{PCR5 rule (Prop. Conflict Redist.)} fusion rule, one gets
$$m_{PCR5}(A)=0.236167\quad m_{PCR5}(B)=0.276500 \quad m_{PCR5}(C)=0.333333$$
$$m_{PCR5}(A\cup B)=0.047500\quad m_{PCR5}(B\cup C)=0.141612$$
\noindent
and if one applies any of BCR12 or BCR17 rules for conditioning the (combined) prior $m_{PCR5}(.)$, one finally gets the same final result with BCR12 and BCR17, i.e.
$$m_{F_{\tiny{PCR5}}C_{\tiny{BCR17}}}(.|B\cup C)=m_{F_{\tiny{PCR5}}C_{\tiny{BCR12}}}(.|B\cup C)$$ \noindent with
\begin{align*}
& m_{F_{\tiny{PCR5}}C_{\tiny{BCR12}}}(B|B\cup C)=0.415159\\
& m_{F_{\tiny{PCR5}}C_{\tiny{BCR12}}}(C|B\cup C)=0.443229\\
& m_{F_{\tiny{PCR5}}C_{\tiny{BCR12}}}(B\cup C|B\cup C)=0.141612\\
\end{align*}
\end{itemize}

From cases 1 and 2, one has proved that there exists at least one example for which
PCR5 fusion and BCR12/17 conditioning rules do not commute since 
$$m_{F_{\tiny{PCR5}}C_{\tiny{BCR12/17}}}(.|B\cup C)\neq m_{C_{\tiny{BCR12/17}}F_{\tiny{PCR5}}}(.|B\cup C).$$

\begin{itemize}
\item Case 3 : BCR12/BCR17-based Conditioning followed by Dempster's rule-based Fusion

If we consider the same masses $m_1(.)$ and $m_2(.)$ and if we apply the BCR12 or BCR17 to each of them, one gets same result, i.e.
$$m_1(B|B\cup C)=0.25   \qquad m_1(C|B\cup C)=0.30 \qquad m_1(B\cup C|B\cup C)=0.45$$
$$m_2(B|B\cup C)=15/35 \qquad m_2(C|B\cup C)=12/35 \qquad m_2(B\cup C|B\cup C)=8/35$$
\noindent
then if one combines them with Dempster’s rule, one finally gets
$$m_{C_{\tiny{BCR12}}F_{\tiny{DS}}}(B|B\cup C)=\frac{125}{275}$$
$$m_{C_{\tiny{BCR12}}F_{\tiny{DS}}}(C|B\cup C)=\frac{114}{275}$$
$$m_{C_{\tiny{BCR12}}F_{\tiny{DS}}}(C|B\cup C)=\frac{36}{275}$$
\noindent
and same result for $m_{C_{\tiny{BCR17}}F_{\tiny{DS}}}(.)$.

\end{itemize}

\begin{itemize}
\item Case 4 : Dempster's rule based Fusion followed by BCR12/BCR17-based Conditioning

If we apply first the fusion of $m_1(.)$ with $m_2(.)$ with Dempster's rule of combination, one gets
$$m_{DS}(A)=10/59\quad m_{DS}(B)= 22/59 \quad m_{DS}(C)=19/59$$
$$m_{DS}(A\cup B)= 2/59 \quad m_{DS}(B\cup C)=6/59$$
\noindent and if one applies BCR12 (or BCR17) for conditioning the prior $m_{DS}(.)$, one finally gets (same result is obtained with BCR17)
$$m_{F_{\tiny{DS}}C_{\tiny{BCR12}}}(B|B\cup C)=\frac{1348}{2773}$$
$$m_{F_{\tiny{DS}}C_{\tiny{BCR12}}}(C|B\cup C)=\frac{1083}{2773}$$
$$m_{F_{\tiny{DS}}C_{\tiny{BCR12}}}(B\cup C|B\cup C)=\frac{342}{2773}$$

\noindent
In BCR12, $m_{DS}(A)=10/59$ is distributed to $B$, $C$, $B\cup C$ proportionally to their masses, i.e.
$$\frac{x_B}{22/59}=\frac{y_C}{19/59}=\frac{z_{B\cup C}}{6/59}=\frac{10/59}{47/59}=\frac{10}{47}$$
\noindent
whence $x_B=(22/59)\cdot (10/47)=220/2773$, $y_C=(19/59)\cdot (10/47)=190/2773$ and $z_{B\cup C}=(6/59)\cdot (10/47)=60/2773$, and $m_{DS}(A\cup B)= 2/59$ is distributed to $B$ only, since $B$ is the $1$-largest.\\

\noindent In BCR17, $m_{DS}(A)=10/59$ is similarly distributed to $B$, $C$, $B\cup C$  and $m_{DS}(A\cup B)$ is also distributed to $B$ only, since $B\subset A$ and $m_{DS}(B)>0$ and $B$ is the only element with such properties. Herein BCR12 and BCR17 give the same result.\\

\end{itemize}

Therefore from cases 3 and 4, we see that BCR12 (and BCR17) don't commute with Demspter's rule for Shafer's model and a non-Bayesian bba since
$$m_{C_{\tiny{BCR12}}F_{\tiny{DS}}}(.|B\cup C)\neq m_{F_{\tiny{DS}}C_{\tiny{BCR12}}}(.|B\cup C).$$

\begin{itemize}
\item Case 5 : SCR-based Conditioning followed by Dempster's rule based Fusion

If we consider the masses $m_1(.)$ and $m_2(.)$ and if we apply the SCR to each of them for conditioning, one gets
$$m_1(B|B\cup C)=2/7 \qquad m_1(C|B\cup C)=2/7 \qquad m_1(B\cup C|B\cup C)=3/7$$
$$m_2(B|B\cup C)=4/9 \qquad m_2(C|B\cup C)=3/9 \qquad m_2(B\cup C|B\cup C)=2/9$$
\noindent
then if one combines them with Dempster’s rule, one finally gets
$$m_{C_{\tiny{SCR}}F_{\tiny{DS}}}(B|B\cup C)=\frac{24}{49}\quad m_{C_{\tiny{SCR}}F_{\tiny{DS}}}(C|B\cup C)=\frac{19}{49}\quad m_{C_{\tiny{SCR}}F_{\tiny{DS}}}(C|B\cup C)=\frac{6}{49}$$

\end{itemize}

\begin{itemize}
\item Case 6 : Dempster's rule based Fusion followed by the SCR-based Conditioning

If we apply first the fusion of $m_1(.)$ with $m_2(.)$ with Dempster's rule of combination, one gets
$$m_{DS}(A)=10/59\quad m_{DS}(B)= 22/59 \quad m_{DS}(C)=19/59$$
$$m_{DS}(A\cup B)= 2/59 \quad m_{DS}(B\cup C)=6/59$$
\noindent and if one applies SCR for conditioning the prior $m_{DS}(.)$, one finally gets
$$m_{F_{\tiny{DS}}C_{\tiny{SCR}}}(B|B\cup C)=\frac{24}{49}\quad m_{F_{\tiny{DS}}C_{\tiny{BCRs}}}(C|B\cup C)=\frac{19}{49} \quad m_{F_{\tiny{DS}}C_{\tiny{BCRs}}}(B\cup C|B\cup C)=\frac{6}{49}$$
\end{itemize}

\noindent From cases 5 and 6, we verify that SCR commutes with Demspter's rule for Shafer's model and non-Bayesian bba\footnote{This property has been proved by Shafer in \cite{Shafer_1976}.} because
$$m_{C_{\tiny{SCR}}F_{\tiny{DS}}}(.|B\cup C)=m_{F_{\tiny{DS}}C_{\tiny{SCR}}}(.|B\cup C).$$

\section{Conclusion}

We have proposed in this paper several new Belief Conditioning Rules (BCRs)\index{BCR (Belief Conditioning Rule)} in order to adjust a given prior bba $m(.)$ with respect to the new conditioning information that have come in.
The BCRs depend on the model of $D^\Theta$. Several examples were presented that compared these BCRs among themselves and as well with Shafer’s Conditioning Rule (SCD). Except for SCD, in general the BCRs do not commute with the fusion rules, and the sequence in which they should be combined depends on the chronology of information received.

\end{document}